\documentclass{article}

\usepackage[preprint]{neurips_2026}
\makeatletter
\renewcommand{\@notice}{}
\makeatother

\usepackage[utf8]{inputenc}
\usepackage[T1]{fontenc}
\usepackage{hyperref}
\usepackage{url}
\usepackage{booktabs}
\usepackage{amsfonts}
\usepackage{nicefrac}
\usepackage{microtype}
\usepackage{xcolor}

\usepackage{amsmath}
\usepackage{amssymb}
\usepackage{graphicx}
\usepackage{multirow}
\usepackage{listings}
\usepackage{inconsolata}
\usepackage{adjustbox}
\usepackage{etoolbox}
\usepackage{placeins}
\usepackage{float}

\BeforeBeginEnvironment{tabular}{\begin{adjustbox}{max width=\linewidth}}
\AfterEndEnvironment{tabular}{\end{adjustbox}}


\newcommand{\sig}[1]{$^{\textrm{#1}}$}
\newcommand{\sigl}[1]{$^{\textrm{#1}}$}

\title{Lost in the Request: How Communication Variation Disrupts Retrieval and Action in Email Agents}

\author{
  Feng Chen, Ritam Dutt, Atnaz Taheri, Alex Williams \\
  \texttt{fengc9@uw.edu}
}

\begin{document}

\maketitle

\begin{abstract} 
An email assistant should not complete less work simply because a user phrases the same request differently. Yet most benchmarks test each task with only one canonical request, leaving this form of robustness largely unmeasured. We test whether email assistants remain reliable when the requested information, available evidence, and expected outcome stay fixed, but the communication style or English variety changes. We construct validated variants along five communication-style axes and four rule-based dialect conditions, and evaluate them on three benchmarks: a retrieval-augmented generation (RAG) pipeline and two tool-using agents. Indirect requests reduce performance on all three benchmarks, while formal requests reduce performance on both agentic benchmarks. Examining the systems more closely shows that these failures have different causes. Verbose requests mainly hurt a lexical retriever by making the relevant email harder to find. By contrast, indirect and dialect variants remain harmful even when the relevant email is retrieved. In the agentic setting, indirect and formal requests mainly cause the agents to omit required actions, not to take more unsupported actions. These results show that a successful response is not enough to establish robustness: evaluations should vary how requests are expressed and separately measure whether agents complete the requested work. 
\end{abstract}

\section{Introduction}
\label{sec:introduction}

An email assistant should not complete less work because a user asks indirectly, writes formally, adds surrounding context, or uses a different variety of English. These differences are common in real communication, but they do not necessarily change what the user wants. If the requested information, relevant evidence, named entities, and intended outcome remain the same, a reliable assistant should return the same answer or carry out the same actions.

Current evaluations rarely test this expectation. Most benchmarks represent each task with one canonical request, even though the same task can be expressed in many ways. Prior work shows that semantically equivalent query formulations can change retrieval results, including for recent retrieval architectures~\citep{penha2022evaluating,hagen2024revisiting}, and that linguistic reformulation affects both retrieval and answer generation in retrieval-augmented question answering~\citep{cao2026out}. Performance also varies across systematic varieties of English~\citep{ziems2022value,ziems2023multi,sap2019risk,srirag2025evaluating}. These findings establish that wording matters for finding and answering information, but they leave an important question for interactive agents: can a change in wording alter whether the requested work is completed? This question matters because an agent can fail without producing an obviously bad response. An assistant may identify the task, call some appropriate tools, and reply fluently while still omitting an action needed to finish the request. Conversely, it may retrieve the wrong evidence or take an unsupported action. These failures require different responses from system designers, but a single end-to-end success score does not distinguish them. Diagnostic agent benchmarks similarly show that aggregate scores can combine failures such as skipping a required tool, ignoring a tool result, or producing unsupported output~\citep{kokane2024toolscan,soni2026toolfailbench}. Robustness evaluation should therefore ask not only whether performance changes, but what the system stopped doing correctly.

We study whether email assistants remain reliable when the same request is expressed in different ways. We evaluate the two settings for email assistants: the \emph{retrieval-augmented generation} (RAG) setting, where the system retrieves evidence from the archive and answers from it, and the \emph{agentic} setting, where the system calls tools to carry out the request. Covering both lets us examine whether communication variation affects the information an assistant returns as well as the actions it performs. We vary communication style and English variety while keeping the underlying task and expected outcome unchanged, and validate the resulting requests to ensure that they preserve the original intent. Our work makes three contributions:

\begin{itemize}
\item \textbf{A framework for generating validated request variants.} Figure~\ref{fig:framework} shows it. An LLM rewriter produces the communication-style poles and a rule-based transducer produces the dialect varieties, and a set of gates admits a variant only once it still requests the information or action asked for, carries the relevant task details, and expects the same outcome.

\item \textbf{An evaluation spanning RAG and agentic settings.} We apply the framework to email question answering and to tool-using agents, so that the same variation can be traced through retrieval, answering, and execution.

\item \textbf{A closer analysis of assistant failures.} We examine not only whether performance changes, but also whether failures come from finding the wrong information, producing an incorrect answer, omitting required actions, or taking unsupported actions.
\end{itemize}
\section{Evaluation Framework}
\label{sec:framework}

\begin{figure*}[t]
\centering
\includegraphics[width=\textwidth]{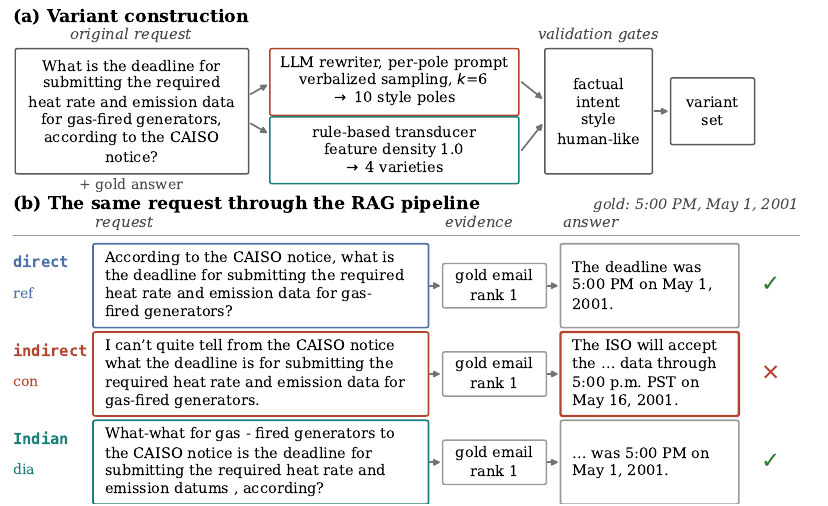}
\caption{Variant construction, and one EnronQA request carried through it. (a) A request is rewritten into ten communication-style poles and four English varieties; a variant enters the evaluation set only after clearing the gates of \S\ref{subsec:validation}. (b) The same request under three variants. The gold email is retrieved at rank 1 in all three cases, but the indirect variant produces an incorrect answer. \textsc{ref}, \textsc{con}, and \textsc{dia} denote reference, contrastive, and dialect conditions, respectively.}
\label{fig:framework}
\end{figure*}

We evaluate whether a system fulfills the same request when the request is expressed differently. We consider two forms of variation. \emph{Communication style} changes how a writer presents a request. \emph{Dialect} changes grammatical and dialectal features according to patterns documented for varieties of English. Figure~\ref{fig:framework} shows how a variant is built and follows one request through a RAG pipeline.

\subsection{Variation Taxonomy}
\label{subsec:taxonomy}

\subsubsection{Communication-Style Axes}
\label{subsubsec:style-taxonomy}

The five axes and the definition of each pole are listed in Table~\ref{tab:style-axes}. Each style axis pairs a \emph{reference} pole, a plain rewrite that carries no marked style, with a \emph{contrastive} pole that carries the property under test. Formality is the exception: neither a formal nor a casual register is neutral, so both of its poles are contrastive. Each axis rests on a documented distinction: directness on the contrast between on-record and off-record requests~\citep{brown1987politeness,mannekote2025making}; formality on register variation~\citep{pavlick2016empirical,rao2018dear}; verbosity on evidence that input length and the position of the relevant information affect model behavior~\citep{levy2024same,liu2024lost}; emotion on evidence that emotional framing changes behavior on an otherwise fixed task~\citep{li2023large,sharma2024towards}; and grammaticality on reported sensitivity to grammatical form~\citep{cao2026out} and on work measuring encoder robustness to grammatical error~\citep{yin2020robustness}. Appendix~\ref{app:axes} describes each axis in full.

\begin{table*}[t]
\centering
\small
\setlength{\tabcolsep}{3pt}
\begin{tabular}{@{}llll@{}}
\toprule
\textbf{Axis} & \textbf{Pole} & & \textbf{Operational definition} \\
\midrule
\multirow{2}{*}{Directness} & \texttt{direct} & ref & The request is stated explicitly as a question or command. \\
 & \texttt{indirect} & con & The request is conveyed through a hint, stated need, uncertainty, or observation. \\
\addlinespace[2pt]
\multirow{2}{*}{Formality} & \texttt{formal} & con & Complete sentences, standard punctuation, no slang or contractions. \\
 & \texttt{casual} & con & Informal wording, contractions, shorthand, lowercase typing, or loose punctuation. \\
\addlinespace[2pt]
\multirow{2}{*}{Verbosity} & \texttt{concise} & ref & The request is stated briefly, with the main ask near the beginning. \\
 & \texttt{verbose} & con & The same request sits within added chatter, hedging, or unrelated context. \\
\addlinespace[2pt]
\multirow{2}{*}{Emotion} & \texttt{calm} & ref & The request is presented in neutral, matter-of-fact language. \\
 & \texttt{charged} & con & The request expresses frustration, stress, anxiety, or urgency. \\
\addlinespace[2pt]
\multirow{2}{*}{Gramm.} & \texttt{grammatical} & ref & Standard written English grammar and word order. \\
 & \texttt{non-gramm.} & con & Restructured word order with systematic grammatical slips. \\
\bottomrule
\end{tabular}
\caption{The five communication-style axes and their poles. \emph{ref} marks a reference pole, a plain rewrite similar to the original query; \emph{con} marks a contrastive pole.}
\label{tab:style-axes}
\end{table*}

\subsubsection{Dialect Conditions}
\label{subsubsec:dialect-taxonomy}

We evaluate four varieties: Indian English, Colloquial Singapore English, Black South African English, and Nigerian English. Dialect differs from the non-grammatical pole in both source and interpretation. The non-grammatical pole is a constructed set of learner-like slips deliberately detached from any speech community, whereas the dialect conditions apply features documented for particular varieties. The variants are synthetic stress tests rather than samples of how individual speakers write, and every claim we make concerns system behavior rather than speakers.

\subsection{Variant Construction}
\label{subsec:construction}

\paragraph{Communication-style rewriting.} Every query is rewritten once into each of the ten poles. Rewriting uses \texttt{gpt-5.4-mini} at temperature $1.0$, with a per-pole prompt that describes the target style and forbids changing fact. For the six contrastive poles we use verbalized sampling \citep{zhang2025verbalized}: one call returns $k{=}6$ candidates, each written to a different framing drawn from a per-pole menu, which is what keeps a pole from collapsing onto similar phrasing. Among candidates clearing every applicable gate described in \ref{subsec:validation}, we keep the one whose opening is least used so far in the run, so that no fixed menu position is systematically favored. The four reference poles are generated as a single plain rewrite with maximum of 3 retries. Full prompts and menus are in Appendix~\ref{app:prompts}.  

\paragraph{Rule-based dialect transformation.} Dialect variants are produced by the transducer of \citet{ziems2023multi},\footnote{\url{https://value-nlp.github.io}} which maps Standard American English into synthetic dialect text by applying sets of grammatical and dialectal features documented for each variety. Generation is rule-based and reproducible under a fixed seed (we use 0), at a feature density $1.0$. Operative literals such as usernames, addresses, times, and event names are masked before transformation and restored afterwards, so they survive byte-identically.

\subsection{Variant Validation}
\label{subsec:validation}

Across a base query and all of its variants we hold fixed the information need or requested action, the supporting evidence, named entities, numbers, dates, protected literals, and the expected outcome. Only the wording may differ. A candidate enters the evaluation set only after clearing every gate that applies to its pole.

A deterministic \emph{factual gate} requires the variant to be non-empty, to differ from the original, and to preserve every numeric token and protected literal. An \emph{intent gate}, a rubric judge (\texttt{gpt-5.4}, temperature $0.0$), asks whether the variant requests exactly the same information. This gate is what protects the paired design: a variant that clears it is still answerable by the same gold answer, so a difference in outcome cannot be a difference in task. A \emph{style gate} asks whether the variant realizes its target pole, given that pole's operational definition from Table~\ref{tab:style-axes}; for dialect it instead requires at least two features documented in the electronic World Atlas of Varieties of English to have changed~\citep{kortmann2013electronic}. A \emph{human-likeness gate} asks whether the text reads as something a person would type, which guards against rewrites that reach the style label while reading synthetic. It applies to four of the six contrastive poles except verbose and non-grammatical (prompts in Appendix~\ref{app:gates}). Two authors independently labeled the same 120 variants, 20 per contrastive pole, answering the questions the gates ask. Agreement was complete on intent ($\kappa{=}1.00$), and reached $82.5\%$ on style and $80.0\%$ on human-likeness.
\section{Experimental Setup}
\label{sec:experiments}

\subsection{Evaluation Tasks}
\label{subsec:tasks}

We apply the framework to three tasks, one RAG task and two agentic benchmarks. 

\paragraph{EnronQA (RAG).} Retrieval-augmented question answering over a personal email archive \citet{ryan2025enronqa}. We sample 400 questions from the training split under an isolated-axes design: four bins of 100, where each complex bin flips exactly one complexity property against the \emph{simple} control. The three complex bins are \emph{3-hop} (three or more reasoning hops), \emph{multi-question} (the question asks for two or more distinct things), and \emph{threaded} (resolution depends on earlier messages in the thread)(detail in Appendix~\ref{app:sampling}). Holding the other two properties at their control setting is what makes the per-bin breakdown of \S\ref{subsubsec:stage-rag} attributable to one property rather than to overall difficulty. Each question is paired with a gold answer and the gold email that contains it, and retrieval runs over the target user's own inbox.

\paragraph{ToolTalk (Agentic).} A multi-turn tool-using benchmark \citep{farn2023tooltalk}. We use the \emph{hard} split: 50 conversations comprising 194 user turns over 28 tool APIs. Grading compares the agent's calls against ground-truth API calls. The harness is teacher-forced: the conversation is rebuilt from the gold trajectory and the databases are reset before each assistant turn, so an error on one turn cannot propagate to the next.

\paragraph{WorkBench (Agentic).} A sandbox benchmark of workplace tasks across five domains~\citep{styles2024workbench}. We use its 90-task email domain, the subset solvable with the email toolkit alone, and expose all 26 tools on every task so that the agent must still select among them. Grading is by database state: a task is correct only if executing the agent's predicted calls leaves the sandbox in the ground-truth state, which makes the metric binary and unforgiving of a single wrong argument. We report task accuracy.

\subsection{Models and Retrieval Systems}
\label{subsec:models}

\paragraph{Retrieval.} EnronQA is run with three retrievers over the same corpus, all returning the top $k{=}5$ emails. BM25 is lexical (\texttt{rank\_bm25}) and indexes the full text of every email~\citep{robertson2009probabilistic}. SPLADE is learned-sparse (\texttt{naver/splade-v3}), producing term-weight vectors scored by dot product~\citep{formal2021splade}. BGE-M3 is dense (\texttt{BAAI/bge-m3})~\citep{chen2024bge}. Both neural retrievers truncate to 512 tokens. Retrieval is identical within a retriever, so any difference between answerers under a fixed retriever is a generation difference. We use SPLADE as the headline retriever because it provides the strongest baseline retrieval among the tested systems.

\paragraph{Answering and grading.} We evaluate three models: \texttt{gpt-5.6-terra}, \texttt{gpt-5.4-mini}, and \texttt{Kimi-K2.6}. All headline results use \texttt{gpt-5.6-terra}, which is run on all three benchmarks. The cross-model comparison of \S\ref{subsec:result-generality} is restricted to EnronQA and ToolTalk: running the full ten-pole grid on WorkBench for the two additional models was beyond the compute available, so that benchmark is reported on the headline model alone and the coverage actually obtained is stated in Table~\ref{app:coverage} so that what is missing is explicit. On EnronQA each model receives the retrieved emails and the question and is instructed to answer in one sentence using only the information in the email. Each agent condition is repeated with $K{=}3$ replicates whose per-scenario metrics are averaged before any test. Answers on EnronQA are graded by \texttt{gpt-5.4} at temperature $0$, which sees the email, the question, the gold answer, and the candidate answer and returns a single-line judgement.

\subsection{Evaluation Metrics}
\label{subsec:metrics}

Each task has its own success metric, so a percentage point is not the same quantity across benchmarks and magnitudes should not be compared between columns. EnronQA reports judged answer accuracy, and additionally Recall@5, the rate at which the gold email appears in the retrieved set. ToolTalk reports conversation success, which requires every ground-truth call to be matched with no incorrect action, along with recall of required calls, precision, and the incorrect-action rate. WorkBench reports task accuracy against the ground-truth sandbox state, and additionally the F1 between predicted and ground-truth tool calls, the rate at which no tool call is issued, and the rate at which execution leaves an unwanted change to the sandbox.

\subsection{Statistical Analysis}
\label{subsec:stats}

Every variant is compared against the base item on the items present in both conditions. Detail shows in Appendix~\ref{app:stats}. For EnronQA, each item yields one binary outcome from a single run, so the paired difference is $-1$, $0$ or $+1$ . We use the exact McNemar test on those pairs, which conditions on their number and is exact at any sample size. For the other two agent benchmarks, each condition is run $K{=}3$ times, so for item $i$ we average a metric over replicates before any test, then form the paired difference $d_i = \bar{v}_i - \bar{n}_i$ against the base item. We test the mean of those differences with a paired percentile bootstrap over items, resampling item indices with replacement $B{=}10{,}000$ times under a fixed seed, reading the 95\% interval off the $2.5$ and $97.5$ percentiles of the resampled means and a two-sided $p$-value off the proportion of them falling either side of zero.
\section{Results}
\label{sec:results}

\subsection{Effects of Style and Dialect Variation}
\label{subsec:result-selectivity}

Table~\ref{tab:main-style} gives the effect of every condition on all three benchmarks, intervals and $p$-values in Table~\ref{app:full-detail}. Indirect phrasing is the only condition that degrades every benchmark. It costs $-3.2$ points of answer accuracy on EnronQA ($p{=}.007$), $-22.4$ points of conversation success on ToolTalk ($p{<}.001$), and $-10.7$ points of task accuracy on WorkBench ($p{=}.002$).

Formality separates the retrieval task from the two agentic ones. Both formality poles are null on EnronQA ($-1.0$ each, $p{=}.39$ and $p{=}.34$), while \texttt{formal} is significant on both agent benchmarks, costing $-13.9$ on ToolTalk and $-12.2$ on WorkBench. The \texttt{casual} pole is not significant on any benchmark, making the formal pole the more consistently harmful in this axis.

Dialect variation degrades every benchmark in the mean, but its strength varies by variety. On EnronQA all four varieties are significant, from $-2.6$ (Singapore) to $-3.9$ (Indian). Singapore is significant on all three benchmarks and Indian on two, whereas Black South African and Nigerian reach significance on one. On EnronQA, the pooled dialect estimate is numerically larger than the estimate for the non-grammatical pole: $-3.0$ versus $-1.8$ percentage points.

The remaining conditions are selective. Verbosity is significant on EnronQA ($-3.0$) alone. Emotionally charged phrasing reaches significance on WorkBench ($-8.5$). Non-grammatical phrasing is significant only on ToolTalk ($-9.3$). The reference poles behave as the control they are intended to be. Across the $12$ reference cells of the headline model, only one reaches $p{<}.05$: \texttt{concise} on ToolTalk ($-9.0$, $p{=}.010$), where shortening an already-terse instruction is not a neutral operation. Input-side similarity (\S\ref{subsec:validation}) shows the same separation between reference and contrastive rewrites before any system is involved.

\begin{table*}[t]
\centering
\small
\setlength{\tabcolsep}{4pt}
\begin{tabular}{ll l rrr}
\toprule
\textbf{Axis} & \textbf{Condition} & & \textbf{EnronQA} & \textbf{ToolTalk} & \textbf{WorkBench} \\
 & & & $\Delta$acc & $\Delta$succ & $\Delta$acc \\
\midrule
\multicolumn{3}{@{}l}{\emph{unperturbed baseline}} & 92.8 & 52.7 & 90.7 \\
\midrule
Directness & \texttt{direct} & ref & +0.3\sig{} & -2.0\sig{} & -2.6\sig{} \\
 & \texttt{indirect} & con & \textbf{-3.2}\sig{**} & \textbf{-22.4}\sig{***} & \textbf{-10.7}\sig{**} \\
\addlinespace[2pt]
Formality & \texttt{formal} & con & -1.0\sig{} & \textbf{-13.9}\sig{***} & \textbf{-12.2}\sig{**} \\
 & \texttt{casual} & con & -1.0\sig{} & -4.0\sig{} & -0.4\sig{} \\
\addlinespace[2pt]
Verbosity & \texttt{concise} & ref & +0.0\sig{} & \textbf{-9.0}\sig{**} & +4.1\sig{} \\
 & \texttt{verbose} & con & \textbf{-3.0}\sig{*} & -4.7\sig{} & -1.9\sig{} \\
\addlinespace[2pt]
Emotion & \texttt{calm} & ref & -0.5\sig{} & +1.3\sig{} & -1.9\sig{} \\
 & \texttt{charged} & con & +0.5\sig{} & -8.7\sig{} & \textbf{-8.5}\sig{*} \\
\addlinespace[2pt]
Grammaticality & \texttt{grammatical} & ref & +0.0\sig{} & -4.8\sig{} & -2.6\sig{} \\
 & \texttt{non-grammatical} & con & -1.8\sig{} & \textbf{-9.3}\sig{*} & -4.1\sig{} \\
\addlinespace[2pt]
Dialect & \texttt{Indian} & con & \textbf{-3.9}\sig{**} & \textbf{-14.7}\sig{**} & -1.9\sig{} \\
 & \texttt{Singapore} & con & \textbf{-2.6}\sig{*} & \textbf{-8.7}\sig{*} & \textbf{-7.0}\sig{*} \\
 & \texttt{Black S. African} & con & \textbf{-2.7}\sig{*} & +0.0\sig{} & -0.7\sig{} \\
 & \texttt{Nigerian} & con & \textbf{-2.7}\sig{*} & -6.8\sig{} & -0.4\sig{} \\
\bottomrule
\end{tabular}
\caption{Unperturbed baseline and paired $\Delta$ per condition, in points on each benchmark's own metric, for the headline model \texttt{gpt-5.6-terra} (EnronQA on SPLADE). \emph{ref} rows are the meaning-preserving reference poles and \emph{con} rows the contrastive poles; formality has no neutral pole, so both of its rows are contrastive. Per-cell $n$, intervals and the test used for each benchmark (\S\ref{subsec:stats}) are in Table~\ref{app:full-detail}. Raw $p$: \sigl{*}$p{<}.05$, \sigl{**}$p{<}.01$, \sigl{***}$p{<}.001$.}
\label{tab:main-style}
\end{table*}

\subsection{Where the Loss Arises}
\label{subsec:result-stages}

\subsubsection{RAG Setting}
\label{subsubsec:stage-rag}

\begin{figure}[t]
\centering
\includegraphics[width=\columnwidth]{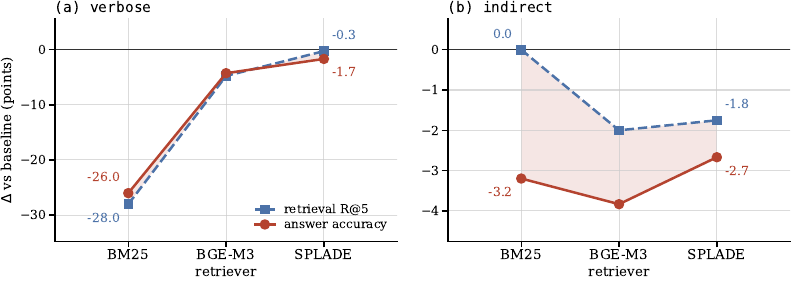}
\caption{Accuracy $\Delta$ and gold-email recall $\Delta$ for verbose and indirect against unperturbed request under three retrievers on EnronQA. Shading marks the gap between the part of the accuracy change not matched by a change in retrieval. Values are averaged over three answerer models; per-model values in Table~\ref{app:grid}.}
\label{fig:stage}
\end{figure}

On EnronQA we can separate retrieval from response, because Recall@5 reports whether the gold email reached the reader at all. Figure~\ref{fig:stage} contrasts two poles across three retrievers.

The verbosity effect turns out to be retriever-dependent. Under lexical BM25, \texttt{verbose} costs $-26.0$ points of accuracy, and Recall@5 falls by $-28.0$: the padding pushes the gold email out of the top five, leading to inaccurate answers. Under learned-sparse SPLADE recall for \texttt{verbose} is close to the baseline ($-0.2$, $p{=}.89$) while accuracy falls only $-3.0$. Under BM25, we further explore the effect of padding length, over the \texttt{verbose} slots whose gold email was reachable at baseline, Recall@5 falls from $95.7\%$ in the least-padded quartile to $44.6\%$ in the most-padded one (Table~\ref{app:verbose-gradient}).

Indirect and dialect phrasing survive the retrieval drop. Under SPLADE, \texttt{indirect} costs $-3.2$ points of accuracy while recall moves $-1.8$, and pooled dialect costs $-3.0$ points of accuracy against $-0.9$ of recall. In both cases the answerer still answers worse, so the loss is downstream of retrieval. Per-pole retrieval effects are in Table~\ref{app:mech-enronqa}, and retrieval quality with per-pole recall intervals are in Table~\ref{app:equiv}.

We labeled the $118$ headline-condition slots (\texttt{gpt-5.6-terra} $\times$ SPLADE) where the gold email was retrieved and the answer was still judged wrong, one code per slot (Appendix~\ref{app:errors}). Unsupported content is the largest category; what separates the rewrite-induced slice is misreading the request rather than declining to answer (Table~\ref{app:error-dist}). When the gold email never reaches it at all, the model answers confidently far more often than reporting the fact absent (Appendix~\ref{app:errors}).

Two of the three complex bins are more fragile than the control. Contrastive rewrites break $4.26\%$ of previously-correct slots on 3-hop questions and $4.08\%$ on multi-question ones, against $1.62\%$ on the simple control and $1.88\%$ on threaded questions. Dialect variation follows the same ordering at a uniformly higher level, from $2.91\%$ on threaded to $5.36\%$ on multi-question (Table~\ref{app:complexity} and Figure~\ref{app:fig-complexity} report how often a rewrite broke a question).

\subsubsection{Agentic Setting}
\label{subsubsec:stage-agents}

\begin{figure*}[t]
\centering
\includegraphics[width=\textwidth]{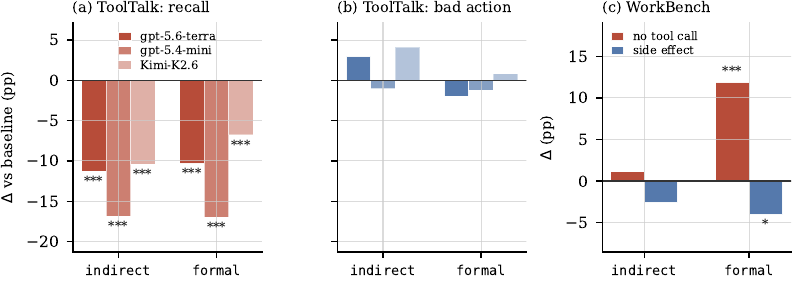}
\caption{Paired $\Delta$ for the indirect and formal poles. Vermillion marks an omission measure and blue an unsupported-action measure; shade marks the model. (a) ToolTalk required-call recall, -17.0 to -6.8~pp over 6 pole$\times$model cells, 6 significant; (b) ToolTalk incorrect-action rate, -1.9 to +4.1~pp, 0 of 6 significant; (a) and (b) share a scale. (c) WorkBench on the headline model, the share of tasks with no tool call beside the share leaving an unwanted change to the sandbox. Per-pole and per-model values in Table~\ref{app:mech-tooltalk} and Table~\ref{app:mech-workbench}. \sigl{*}$p{<}.05$, \sigl{**}$p{<}.01$, \sigl{***}$p{<}.001$.}
\label{fig:omission}
\end{figure*}

The agents can have two failure modes: the agent could take wrong actions, or it could take too few. Figure~\ref{fig:omission} separates them. On ToolTalk the effect is concentrated in a failure to complete required work. Required-call recall falls by $-11.28$ points under \texttt{indirect}, and $-10.32$ under \texttt{formal} (both $p{<}.001$). The incorrect-action rate moves in neither pole: $+2.96$ ($p{=}.14$) and $-1.94$ ($p{=}.35$).

WorkBench shows the similar mechanism (Figure~\ref{fig:omission}c; Table~\ref{app:mech-workbench}). Under \texttt{formal} the share of tasks with no tool calls rises from $0.4\%$ to $12.2\%$ ($p{<}.001$), while unwanted side effects fall from $8.9\%$ to $4.8\%$ ($p{=}.027$). Across both agentic benchmarks the poles that reduce task success reduce the amount of required work performed while leaving the rate of unsupported actions unchanged.

\subsection{Generality Across Models and Systems}
\label{subsec:result-generality}

\begin{table*}[t]
\centering
\small
\setlength{\tabcolsep}{4pt}
\begin{tabular}{l rrr rrr}
\toprule
& \multicolumn{3}{c}{\textbf{EnronQA}} & \multicolumn{3}{c}{\textbf{ToolTalk}} \\
\cmidrule(lr){2-4}\cmidrule(lr){5-7}
\textbf{Pole} & terra & mini & Kimi & terra & mini & Kimi \\
\midrule
\emph{baseline} & 92.8 & 91.2 & 90.2 & 52.7 & 37.3 & 39.3 \\
\midrule
\texttt{indirect} & \textbf{-3.2}\sig{**} & \textbf{-3.2}\sig{*} & -1.5\sig{} & \textbf{-22.4}\sig{***} & \textbf{-19.7}\sig{**} & \textbf{-14.3}\sig{**} \\
\texttt{formal} & -1.0\sig{} & -1.0\sig{} & +0.5\sig{} & \textbf{-13.9}\sig{***} & \textbf{-16.7}\sig{**} & -3.5\sig{} \\
\texttt{casual} & -1.0\sig{} & -1.5\sig{} & -1.2\sig{} & -4.0\sig{} & -6.0\sig{} & -3.3\sig{} \\
\texttt{verbose} & \textbf{-3.0}\sig{*} & -1.5\sig{} & -0.5\sig{} & -4.7\sig{} & +2.7\sig{} & -5.3\sig{} \\
\texttt{charged} & +0.5\sig{} & -1.0\sig{} & -0.3\sig{} & -8.7\sig{} & +3.3\sig{} & -4.7\sig{} \\
\texttt{non-grammatical} & -1.8\sig{} & -2.5\sig{} & +1.0\sig{} & \textbf{-9.3}\sig{*} & \textbf{-17.3}\sig{**} & -9.3\sig{} \\
\bottomrule
\end{tabular}
\caption{Unperturbed baseline and paired $\Delta$ per contrastive pole for three models. The first row is the unperturbed score in percent on each benchmark's own metric; every row below is the paired $\Delta$ in percentage points for one contrastive pole. Restricted to the two benchmarks where all three models were run; the headline model's results on all three are in Table~\ref{tab:main-style} and the coverage obtained is in Table~\ref{app:coverage}. Dialect is in Table~\ref{app:dialect-models}. \sigl{*}$p{<}.05$, \sigl{**}$p{<}.01$, \sigl{***}$p{<}.001$.}
\label{tab:generality}
\end{table*}

Table~\ref{tab:generality} shows every contrastive pole for three models on the two systems where all three models were run (model coverage shown in Table~\ref{app:coverage}). Indirect phrasing is the one pole that generalizes: it is negative in all six model-by-system cells and significant in five, while \texttt{formal} and the non-grammatical pole are each significant in two cells, both on the agent. Higher baseline performance does not imply greater fairness in our experiments. \texttt{gpt-5.6-terra} has the highest unperturbed baseline on both systems, yet on the same twelve contrastive style cells and eight dialect cells it is significant in $5$ and $6$, against $4$ and $4$ for \texttt{gpt-5.4-mini} and $1$ and $0$ for \texttt{Kimi-K2.6}.
\section{Discussion}
\label{sec:discussion}

\subsection{Request Explicitness Affects Performance}
\label{subsec:disc-force}

The most consistent communication-style effect is not the one that changes the most text, but hiding the main request in the query. Parsing every variant against its own original with a dependency parser~\citep{honnibal2020spacy} (Table~\ref{app:verb-fate}), the indirect pole retains the head verb of the request as the main predicate least often on all three benchmarks, from $12.5\%$ of slots on EnronQA down to $4.7\%$ on WorkBench, and the formal pole is in the bottom three poles on all three. The two poles reach that position differently: the indirect pole deletes the original head verb in $16\%$ to $40\%$ of slots, whereas the formal pole retains it in $68\%$ to $91\%$ of slots but frequently places it in a subordinate clause. This pattern suggests that performance is influenced by changing how explicitly the request is expressed~\citep{austin1975things,searle1975indirect, mannekote2025making}. Recent evaluations report that models remain weaker at recovering indirect illocutionary force than at interpreting explicit requests~\citep{koo2025evaluating,solidjonov2026pragmatic}.

\subsection{Incomplete Fulfillment as an Agentic Failure Mode}
\label{subsec:disc-omission}

An end-to-end success score combines two failures with different consequences: an agent can take an unwanted action, or it can omit an action that the request requires. For the indirect and formal poles, the dominant change is in the second type. Shown in both agentic settings, the affected agents complete less of the requested work without a corresponding increase in unsupported actions.

The WorkBench trajectories show two routes into that outcome (Table~\ref{app:mech-workbench}). Under \texttt{formal} the agent more often never starts: the share of tasks with no tool call at all rises from $0.4\%$ to $12.2\%$ ($p{<}.001$), close to the $-12.2$ point accuracy loss. Under \texttt{indirect} that share moves only to $1.5\%$ ($p{=}.25$), while the F1 between predicted and ground-truth calls falls $-8.42$ points ($p{<}.001$), so the agent acts but the calls land in the wrong state. This result complements work showing that aggregate success rates do not identify which tool-use failure occurred~\citep{kokane2024toolscan, soni2026toolfailbench}. We hold the task and tool environment fixed and find that wording alone moves a system between completing a task and skipping required calls.

\subsection{Dialect Robustness and Fair Access}
\label{subsec:disc-dialect}

Dialect variation reduces performance on all three benchmarks. Our finding of performance drops across English varieties is consistent with the disparities reported for language understanding~\citep{ziems2022value,ziems2023multi,srirag2025evaluating}, for social tasks such as hate-speech detection~\citep{sap2019risk}, and for model judgments about people~\citep{hofmann2024dialect}. The dialect transformations apply grammatical features documented for each variety, whereas the non-grammatical pole is a constructed set of slips produced by a language model, which suggests that variation beyond isolated grammatical slips carries part of the effect. It is worth noticing that dialect variants are generated at feature density $1.0$, the attested upper bound, so this condition is a deliberate worst case.

\subsection{Implications for Evaluation and System Design}
\label{subsec:disc-implications}

Three implications can be derived from the results. First, a task should not be represented by a single canonical wording. Retrieval research has shown that results move under semantically equivalent query variation~\citep{penha2022evaluating,hagen2024revisiting}, and \citet{cao2026out} reports the same fragility in the RAG setting. Our results extend that requirement to the agentic setting, where the same variation changes whether the requested work is carried out.

Second, robustness evaluation should be stage-aware, because the same drop in a success score has different causes and therefore different fixes. Replacing BM25 with SPLADE removes most of the verbosity penalty at a comparable baseline. The indirect and dialect penalties persist when the gold evidence is retrieved. Component-level measurement has to come before the choice of mitigation, and choosing a stronger model is not a substitute for it: the model with the highest baseline on both systems where all three models were run is also the one whose results move significantly in the most conditions.

Third, completeness and unsupported action should be reported separately. Required-call recall, failure to act, and read-without-write behavior are computable from a trajectory without a gold final answer, so they can be monitored where end-state grading is unavailable. They are not a substitute for final grading, but they expose a failure that response quality and safety metrics miss, and they distinguish an agent that has become appropriately cautious from one that is quietly doing less of the requested work.
\section{Conclusion} 
\label{sec:conclusion} 
We tested whether RAG-based and agentic assistants fulfill the same task when only the request's communication style or English variety changes. Indirect requests produced the most consistent style loss; verbosity primarily disrupted lexical retrieval; and dialect sensitivity remained after accounting for whether the gold email was retrieved. Across the agentic benchmarks, indirect and formal requests mainly reduced completion of required actions. Evaluations should therefore vary request formulation, inspect intermediate retrieval and execution signals, and report required-action completion separately from unsupported action.
\section*{Limitations}

There are several limitations in this work. First, dialect variants are generated at feature density $1.0$ only, a deliberate worst case rather than an estimate of typical use. Second, the WorkBench \texttt{indirect} pole passes only $64$ of $90$ slots, leading to lower coverage. Third, on ToolTalk the \texttt{concise} reference pole moves $-9.0$ points, so the reference poles are an imperfect control there. Besides, the three benchmarks use different metrics at different operating points, so cross-benchmark comparison is qualitative, and the cross-model comparison covers two of the three because running the full grid on WorkBench was beyond the available compute. Finally, no mitigation approach is tested, and establishing which intervention offers the best trade-off is left to future work.

% \section*{Acknowledgments} 
% we can delete this section...

\FloatBarrier
\bibliographystyle{plainnat}
\bibliography{custom}

\FloatBarrier
\clearpage

\raggedbottom
\preto\section{\FloatBarrier}
\appendix

\section{Communication-Style Axes in Detail}
\label{app:axes}

The five axes summarized in Table~\ref{tab:style-axes} are described here in full.

\paragraph{Directness.} This axis separates an explicit request from one whose information need must be inferred. At the direct pole the request is stated as a question or command; at the indirect pole it is conveyed through a need, hint, observation, or expression of uncertainty. The contrast follows the distinction between on-record and off-record requests in \citet{brown1987politeness} and \citet{mannekote2025making}.

\paragraph{Formality.} This axis separates a polished professional register from a casual, chat-like one, following the register variation studied in \citet{pavlick2016empirical} and \citet{rao2018dear}. Formality has been reported as a source of variation in retrieval-augmented question answering in \citet{cao2026out}.

\paragraph{Verbosity.} This axis separates a concise request from the same request embedded in additional text. The added material is generic chatter, hedging, or digression, and introduces no new fact about the task, the evidence, or the expected answer. Prior work suggests that input length and the position of the relevant information affect model behavior \citep{levy2024same, liu2024lost}.

\paragraph{Emotion.} This axis separates neutral wording from wording that expresses frustration, stress, anxiety, or urgency. Emotional content in a prompt is known to change model behavior on an otherwise fixed task \citep{li2023large}, and models have been shown to revise responses under user pressure \citep{sharma2024towards}. The axis tests whether that occurs when the task, evidence, and expected outcome are held constant.

\paragraph{Grammaticality.} This axis separates standard written English from understandable English carrying systematic grammatical deviation, including the topic or object fronted, the question word moved with the auxiliary dropped, the verb flattened with an adverb carrying the time, or clauses split and reordered. Systematic slips sit on top of that restructuring, including dropped or wrong articles, missing plural marking, and tense or agreement errors. The pole is defined by properties of the text alone and is not conditioned on any writer attribute.

\section{EnronQA Sampling Design}
\label{app:sampling}

EnronQA is the only benchmark in the study for which we choose the items, so the choice is described in full. The goal was a sample on which question complexity could be read as an effect of one property at a time rather than of overall difficulty.

\paragraph{Classification.} Three complexity properties are assigned by a rule-based classifier over the question text alone: reasoning depth (\texttt{1-hop}, \texttt{2-hop}, \texttt{3+-hop}), scored from the presence of a context prefix, a relative clause, a nested reference to something mentioned elsewhere, and two or more clause connectives; question multiplicity (\texttt{single-question}, \texttt{multi-question}), triggered by coordinating \emph{and}, \emph{as well as} or \emph{along with} that introduce a new wh-clause; and context dependency (\texttt{self-contained}, \texttt{thread-dependent}), triggered by explicit reference to an earlier or prior message. Broad threading markers such as \emph{forwarded} or \emph{re:} were deliberately excluded, since a message in a thread can still be self-answerable. A fourth label, reasoning type, is assigned from the question and gold answer under a fixed priority order: causal, comparative, aggregation, temporal, bridging, and otherwise extraction.

\paragraph{Isolation.} Each bin of $100$ requires exactly one of the three properties at its hard setting and the other two at the easy setting. The \emph{simple} control is at most two hops, single-question and self-contained; \emph{3-hop} raises depth only; \emph{multi-question} raises multiplicity only; \emph{threaded} raises context dependency only. The final selection asserts that the number of hard properties is $0$ for every simple item and exactly $1$ for every other item. A difference between the control and one complex bin is therefore a difference in that one property.

\paragraph{Selection.} The pool is the EnronQA train split restricted to users with at least $25$ emails, deduplicated on question text and thinned to at most one question per email. Within each bin, items are drawn stratified by reasoning type so that every type reaches at least $30$ items across the $400$, with the residual filled to $100$ per bin. Two filters run during selection. A near-duplicate guard rejects a candidate whose embedding has cosine similarity $\geq 0.80$ to any question already kept. A content screen runs the question through an oracle answerer and drops it if the provider blocks or the model refuses.

\paragraph{Composition.} The result is $400$ questions over $400$ distinct emails and $99$ users, $100$ per bin. Reasoning types are distributed as extraction $126$, bridging $74$, causal $61$, temporal $57$, aggregation $45$ and comparative $37$. The distribution tries to replicate the distribution in the whole dataset. Retrieval runs over the target user's own inbox, which ranges from $41$ to $1023$ emails with a median of $273$.

\section{Verbalized Sampling}
\label{app:vs}

A single rewrite request per item produces low-diversity output: the model converges on one stock opener and reuses it across items, so the resulting variant set measures one phrasing rather than a style. We therefore generate each contrastive pole with verbalized sampling, in which one call returns $k$ candidates and each candidate is assigned a different framing drawn from a per-pole menu.

\paragraph{Procedure.} For an item with identifier $q$ and a pole with menu $M$, we request $k=6$ candidates in a single call. The candidates are returned as a JSON array. The six rewrites restructure the sentence six different ways rather than repeating one edit. The rotation is a deterministic function of $q$, seeded by an MD5 digest, so runs are reproducible and order-independent.

\paragraph{Selection.} Candidates are first filtered by the deterministic checks of Appendix \ref{app:gates}. The surviving candidates are then scored by the novelty of their opening words, measured as the number of times that opener has already been kept during the run, and the least-used opener wins. Ties are broken by a per-item seeded permutation of the candidate order, which prevents a fixed menu position from being systematically preferred whenever several candidates score equally. The winning candidate is then passed to the judges. If none passes, the call is repeated, twice for most poles and three times for the charged pole.

\paragraph{Prompt template.} The user message is constructed as shown in Appendix \ref{app:vs-template}, where \texttt{k} is the candidate count.

\section{Communication-Style Prompts}
\label{app:prompts}

Each contrastive pole has one system prompt and one menu of framings listed below. The four reference poles use the plain rewrite prompt of Appendix \ref{app:sys-ref} with the per-pole instruction of Appendix \ref{app:ref-instructions}.The rewrite instruction is given to the generator, and the fidelity criterion is supplied to Gate B, the style-realization gate.

\subsection{System Prompt: \texttt{indirect}}
\label{app:sys-indirect}
\begin{lstlisting}[breaklines=true]
You rewrite a user's QUESTION about an email into an INDIRECT, off-record request, changing ONLY its directness — never its meaning.

Hard rules:
- Preserve the information need EXACTLY: a reader must recover the SAME single answer, and it must be answerable by that same answer. Keep every specific detail (names, dates, numbers, quoted phrases, the subject). Add no new fact and do NOT answer it.
- Be genuinely indirect: imply the request via the given mechanism instead of a blunt on-record 'what/who/when' question or command. The output need not be a question.
- Stay specific: the reader must still know exactly which detail is asked about.

Make it read like a REAL person actually typing this, not an assistant:
- First person, natural contractions (I'm, can't, don't, it's, there's).
- Plain, concrete, everyday words — NOT formal, elaborate, or corporate vocabulary, and no nominalisations ('clarification is sought').
- Keep it short and a little uneven; a trailing thought or a sentence fragment is fine.
- Drop assistant scaffolding: no 'I hope this helps', 'kindly', 'I would be grateful', no exhaustive hedging, no stiff register.
- One clear ask; don't over-explain or pile on qualifiers.

Output ONLY the rewritten text: no quotes, labels, or commentary.
\end{lstlisting}

\subsection{Indirectness Mechanisms}
\label{app:menu-indirect}
\begin{lstlisting}[breaklines=true]
1. PREPARATORY (hearer's knowledge): imply the ask by wondering whether the reader happens to know or could tell you the specific thing -- not a bare 'what/who/when' question.
2. DESIRE (speaker's want): state that you would like or want to know the specific thing, as a wish rather than a question.
3. REASON / MOTIVATION: give the reason or goal that makes the specific answer necessary, and let the request follow from it rather than asking outright.
4. UNMET PRECONDITION: note that you still lack the specific piece of information (you don't have it, or can't find where it is stated), so the need for it is implied.
5. EPISTEMIC UNCERTAINTY: say you can't tell or work the specific answer out from the email, leaving the question implied.
6. ASSOCIATIVE HINT: name the exact detail in the email the answer hangs on and flag that its meaning or reference is unclear, so the reader infers the ask.
7. UNDERSTATEMENT: be tentative and under-specify -- say you're a little fuzzy or hazy on the specific point, without naming it as a question.
8. PRESUPPOSITION: presuppose the specific answer exists in the email and comment around it rather than requesting it outright.
9. HYPOTHETICAL / CONDITIONAL: embedding the request inside the condition.
10. REPORTED / THIRD-PARTY: frame it as someone else having asked you this and you being unable to answer it from the email.
11. GAP OBSERVATION: point out that the email leaves the specific point unspecified or open, implying you want it resolved.
12. HEDGED APOLOGY: preface with a hedge or apology for possibly missing it, then gesture at the specific point.
\end{lstlisting}

\subsection{System Prompt: \texttt{formal}}
\label{app:sys-formal}
\begin{lstlisting}[breaklines=true]
You rewrite a user's QUESTION about an email into a formal, professional register — the way a real, competent professional actually writes a courteous work email — changing ONLY the register, never the meaning.

Hard rules:
- Formal and polite: complete sentences, standard capitalisation and punctuation, no contractions (write 'I would', 'could you', not 'I'd', 'you're') and no slang or abbreviations.
- But sound like a REAL PERSON, not an institution. AVOID robotic officialese: no nominalisation pile-ups ('clarification is sought regarding'), no 'the undersigned', no 'kindly ascertain', no archaic or bureaucratic formulas, no stiff passive throat-clearing.
- A natural professional voice: 'Could you please confirm…', 'I wanted to check…', 'Would you be able to let me know…', 'I am trying to confirm…'.
- One clear, courteous ask. Keep the exact same ask and every specific detail (names, dates, numbers, quoted phrases, the subject); add no new fact and do NOT answer it.

Vary the courteous opening from item to item rather than reusing one stock formula. Output ONLY the rewritten question: no quotes, labels, or commentary.
\end{lstlisting}

\subsection{Formal Framings}
\label{app:menu-formal}
\begin{lstlisting}[breaklines=true]
1. a courteous, direct request — 'Could you please confirm…' / 'I wanted to check…'
2. a polite request that briefly references the email, then asks
3. a professional wanting to verify one specific detail before proceeding
4. a measured, respectful request for clarification
5. a warm but businesslike request from a colleague
6. a professional restating what they need, to be sure they have it right
7. a brief, courteous request that gets to the point politely
8. a considerate request that acknowledges the reader's time
\end{lstlisting}

\subsection{System Prompt: \texttt{casual}}
\label{app:sys-casual}
\begin{lstlisting}[breaklines=true]
You rewrite a user's QUESTION about an email as a casual, chat-like message — the way you'd quickly text or Slack a coworker you're friendly with — changing ONLY the register, never the meaning.

Hard rules:
- Write EVERYTHING in lowercase — no capital letters anywhere, not even at the start of a message, on 'i', or on names (jim, socal, the qcom deal); that all-lowercase look is just how people dash off a quick text.
- Make it genuinely informal, not a formal sentence with a 'hey' bolted on the front. Actually rewrite it into how you'd text: contractions and common short forms (u, ur, pls, thx, lmk, btw, tbh, rn, asap, idk, fyi), everyday chatty wording and light slang (gonna, wanna, kinda, gotta, ngl, lowkey, real quick, any chance), loose or dropped punctuation, and — where it fits — an emoji or two. Lean into the colloquial voice: use several of these touches per message and vary them across rewrites, but keep it readable and human — not a keyboard-smash of slang.
- By DEFAULT strip the stiff, contentless scaffolding a texter wouldn't bother typing — 'according to the email', 'as mentioned in the email', 'in the email chain', 'respectively', 'per your message' — instead of carrying it over from the original. Especially drop a bare 'according to the email' wrapper; keep only a real disambiguator that rides along with it (a topic like 'about option settings in a structured note', a person, or a document name). It's the same ask, just stripped of the formal packaging.
- But KEEP every real detail: names, dates, numbers, quoted titles or phrases, and exactly what's being asked. If an attribution names a specific person or document that pins down which answer is meant, keep that name. Add no new fact and do NOT answer it.

Make it read like a REAL person actually typing this, not an assistant:
- First person, natural contractions (I'm, can't, don't, it's, there's).
- Plain, concrete, everyday words — NOT formal, elaborate, or corporate vocabulary, and no nominalisations ('clarification is sought').
- Keep it short and a little uneven; a trailing thought or a sentence fragment is fine.
- Drop assistant scaffolding: no 'I hope this helps', 'kindly', 'I would be grateful', no exhaustive hedging, no stiff register.
- One clear ask; don't over-explain or pile on qualifiers.

Vary the opening — sometimes a quick 'hey', sometimes just the ask with no greeting at all; don't reuse one stock opener. Output ONLY the rewritten question: no quotes, labels, or commentary.
\end{lstlisting}

\subsection{Casual Framings}
\label{app:menu-casual}
\begin{lstlisting}[breaklines=true]
1. a quick slack ping to a work friend — lowercase, a contraction or two
2. a text dashed off between meetings — no greeting, straight to the ask, loose punctuation
3. a chatty message using a short form or two (lmk, u, pls) and maybe one emoji
4. a breezy 'sry to bug u' ask that trims the formal bits
5. an offhand one-liner, right to the point, no opener
6. a relaxed ping like texting a teammate you get along with
7. a slightly rushed 'real quick' ask with a contraction or two
8. a friendly note that drops the stiff 'according to the email' framing and just asks
\end{lstlisting}

\subsection{System Prompt: \texttt{verbose}}
\label{app:sys-verbose}
\begin{lstlisting}[breaklines=true]
You rewrite a user's QUESTION about an email as a long, rambling message — the way a real, chatty person over-writes — changing ONLY the length and packaging, never the meaning.

Hard rules:
- Make it clearly LONG and wandering, and BURY the actual question in the MIDDLE — never state the ask cleanly up front or as a tidy final line.
- Pad it like a real person rambles: a concrete aside, thinking out loud, a specific tangent, an afterthought, a run-on. NOT with abstract self-commentary about being thorough / precise / careful, and NOT by stacking hedges and qualifiers — that reads like an AI.
- The padding must be CONTENT-FREE about the email: ramble about yourself, your day, your workload, your process — never invent any new fact, detail, name, number, or claim about the email itself.
- Keep the exact same ask and every specific detail (names, dates, numbers, quoted phrases, the subject); do NOT answer it.

Write in first person, natural contractions, plain concrete words. Vary the rambling and — importantly — the OPENING WORDS from item to item: do NOT keep starting with 'I keep', 'I've been', or the same first few words; lead sometimes with the aside, sometimes mid-thought, sometimes with a complaint or an apology. Output ONLY the rewritten message: no quotes, labels, or commentary.
\end{lstlisting}

\subsection{Verbose Framings}
\label{app:menu-verbose}
\begin{lstlisting}[breaklines=true]
1. open with a concrete, unrelated aside about your day or workload (a meeting you just left, an overflowing inbox, needing coffee) before drifting toward the ask
2. open by apologising for the rambly message, then wander through an afterthought or two before the ask lands mid-paragraph
3. go off on one specific tangent about why you need this right now, then slip the ask in partway through
4. start mid-thought as if continuing an earlier conversation, backtrack and correct yourself, then tuck the ask in among the afterthoughts
5. mildly complain about how much email or paperwork you're wading through today, then get to the question in the middle of it
6. let several loosely-related thoughts pile up in one long run-on so the ask sits somewhere in the middle
7. open with a small worry about getting a detail wrong, ramble around it, and bury the ask inside the reassurance you give yourself
8. narrate yourself hunting around for this — scrolling, losing your place — before and after you finally ask
\end{lstlisting}

\subsection{System Prompt: \texttt{charged}}
\label{app:sys-charged}
\begin{lstlisting}[breaklines=true]
You ARE a real person under genuine stress, dashing off a quick question about an email. Your exact situation is the persona you're given. Write the question the way THAT person, in THAT moment, would actually type it — changing ONLY the emotional tone, never the meaning.

Hard rules:
- Stay in character and let the pressure of your situation bleed into the whole message — word choice, rhythm, what you fixate on — so the frustration, stress, anxiety, or urgency is UNMISTAKABLE. It must NOT read as a calm, neutral question.
- Weave the emotion INTO the ask itself, and RETYPE the whole question in your own rushed words -- keep every name, date and number, but the wording around them must be YOURS, not the original formal sentence pasted back. Do NOT open with a detached reaction and then paste the tidy question after it (NOT 'I'm asking again -- what two programs...', NOT 'Before this lands on me, what's the context...'), and do NOT tack an emotional bolt-on ('now.', 'ASAP!!') onto an otherwise neutral sentence -- the strain has to run THROUGH how you ask, from the first word to the last.
- Show it the way people actually type under pressure, don't announce it: terse, clipped phrasing, fixating on the one detail, an exasperated or worried aside, a pointed jab, and AT MOST one exclamation mark or one capitalised word. Do NOT stack intensifiers ('really so incredibly urgent!!!'), do NOT open with a stock label ('URGENT:', 'I'm so frustrated:', 'Ugh,'), and do NOT sound like an assistant.
- Your stress is your own state of mind, not a new fact about the email. Keep the exact same ask and add no new factual detail; the emotional wording does NOT count as a new fact.

Make it read like a REAL person actually typing this, not an assistant:
- First person, natural contractions (I'm, can't, don't, it's, there's).
- Plain, concrete, everyday words — NOT formal, elaborate, or corporate vocabulary, and no nominalisations ('clarification is sought').
- Keep it short and a little uneven; a trailing thought or a sentence fragment is fine.
- Drop assistant scaffolding: no 'I hope this helps', 'kindly', 'I would be grateful', no exhaustive hedging, no stiff register.
- One clear ask; don't over-explain or pile on qualifiers.

Output ONLY the rewritten question: no quotes, labels, or commentary.
\end{lstlisting}

\subsection{Emotional-Charge Situations}
\label{app:menu-charged}
\begin{lstlisting}[breaklines=true]
1. you've been heads-down for 12+ hours and you're exhausted and out of patience — you just want this one detail so you can finally log off
2. you're minutes from a hard deadline and still missing this — no time for niceties, you need it right now
3. something's blowing up and your boss is on your back for this exact detail; you're anxious and scrambling
4. you'll be the one blamed if this detail is wrong, and that's stressing you out — you need to be sure before it lands on you
5. you already asked about this once (maybe twice) and heard nothing back; you're fed up and losing patience
6. you're drowning in too many things at once and frazzled — you can barely spare the seconds, so it comes out sharp and clipped
7. a client or an exec is waiting on you and breathing down your neck; the pressure is making you tense and urgent
8. it's late, you're worn down and a bit desperate, quietly dreading what happens if you can't pin this detail down
\end{lstlisting}

\subsection{System Prompt: \texttt{non-grammatical}}
\label{app:sys-nongrammatical}
\begin{lstlisting}[breaklines=true]
You are writing English as a second language: fluent enough to be understood, but you BUILD THE SENTENCE with non-native word order and make real, systematic learner mistakes. Write the request the way YOU would actually type it in a hurry, from scratch — NOT a correct English sentence with a couple of mistakes sprinkled on top.

Hard rules:
- RESTRUCTURE THE SENTENCE — this is the most important rule. The word order and clause structure MUST visibly change from the original; do NOT leave the original sentence intact. Use non-native structural moves such as: front the topic or object ('this report, when you send?', 'the two addresses, what they are?'); put the question word in the middle or at the END instead of the front ('the confirmation come by what two method?', 'the addresses are what?'); drop the question auxiliary do/does/did/will ('why he not tell me?', 'when they arrive?'); flatten verbs to base form and let an adverb carry the time ('yesterday he send it'); or reorder / split / run together the clauses. At least ONE clear structural change is REQUIRED in every rewrite.
- ON TOP of the reorder, add a few systematic grammar and spelling slips: dropped or wrong articles ('a/the'), wrong or missing plural -s, wrong verb tense or agreement, a missing or extra auxiliary, a dropped copula ('is/are'), wrong prepositions ('depend of', 'discuss about'), and occasional misspellings of ORDINARY words ('recieve', 'adress', 'seperate', 'there/their'). These are the SEASONING; the restructured word order is the main dish — a sentence that keeps native word order and only drops an article or adds a typo does NOT count.
- Be UNEVEN, like a real learner: a few errors, some parts fluent. Do NOT stack every possible error at once — it must stay readable and human, NOT gibberish or machine-translation noise.
- Do NOT cheat by leaving the original question perfect and bolting an extra awkward sentence onto the end — the CORE question itself must be restructured.
- Do NOT use any foreign greeting, discourse particle, honorific, accent, or code-switched word (no 'Bonjour', 'Namaste', 'na?', '-san', 'ji', and the like). This is plain English with non-native STRUCTURE and errors, nothing cultural bolted on.
- Keep every name, number, date, quoted title, and specific detail EXACTLY correct (spelling errors are ONLY for ordinary words), keep the exact same ask (answerable by the same answer), and add no new fact about the email.

Write in first person, plain and short. Do NOT begin with any label (never 'Non-native:', 'ESL:', '[L2]' or similar) — start directly with your message. When several rewrites are requested, VARY the structure — each rewrite reorders the sentence a DIFFERENT way and leans on a different mix of mistakes, so the set does not read as one template. Output ONLY the rewritten request: no quotes, labels, or commentary.
\end{lstlisting}

\subsection{Non-Grammatical Structural Profiles}
\label{app:menu-nn}
\begin{lstlisting}[breaklines=true]
1. Topic-fronting — move the object or topic to the FRONT and comment after it, dropping articles and 'is/are' ('this report, already sent?', 'the two addresses, what they are?'). Lead with the reordering; a dropped article or plural may tag along.
2. Question-inversion — break the normal question order: put the wh-word in the middle or at the END and drop the do/does/did/will auxiliary ('Larry Campbell receive confirmation by what two method?', 'the timing and start date is what?', 'why not you told me?'). Lead with the scrambled question structure.
3. Word-order and tense — reorder subject / verb / object and use the wrong or base-form verb with the time set by an adverb, not the verb ('yesterday he send the report', 'next week she call you, why she not say?'). Lead with the word-order shift plus tense slips.
4. Clause-reorder — swap or split the main and subordinate clauses, move the 'according to ...' or 'in the email ...' part to an odd spot, or fuse clauses together ('in the email about the forecast, five issues AA&Co review, the treatment is what?'). Lead with clause restructuring; a preposition slip may come along.
5. Fragment and run-on — chop the sentence into short choppy fragments or fuse it into one long run-on, misplace a modifier, and drop the connecting words ('the post date, the time, for All Shippers Letter #01-317 — what is?'). Lead with the broken flow.
6. Spelling-and-rebuild — rebuild the sentence in a DIFFERENT word order AND misspell ordinary words or mix homophones ('recieve', 'adress', 'seperate', 'there/their') — but keep every name, place, date, number, and quoted title spelled exactly right. The reorder is required; the misspellings sit on top.
\end{lstlisting}

\subsection{Reference-Pole Rewrite System Prompt}
\label{app:sys-ref}
\begin{lstlisting}[breaklines=true]
You rewrite a user's QUESTION about an email to change ONLY its communication STYLE, never its meaning.

Hard rules:
- Preserve the intent EXACTLY: the rewritten question must request the same information and be answerable by the same answer as the original.
- Keep every specific detail from the original: names, dates, numbers, quoted phrases, and the subject of the question.
- Do NOT add any new question, requirement, fact, or assumption, and do NOT answer the question.
- Change ONLY the target style described by the user. Never describe or alter the email itself.
- Output ONLY the rewritten question text — no quotes, no labels, no commentary.
\end{lstlisting}

\subsection{Reference-Pole Rewrite Instructions}
\label{app:ref-instructions}
Given to the generator for the four reference poles, which are produced by a single plain rewrite without verbalized sampling. The contrastive poles are generated from the pole-specific system prompts above instead.
\begin{lstlisting}[breaklines=true]
[directness / direct]
  rewrite_instruction: Rewrite as a fully explicit, on-record request: state exactly what is being asked, up front, as a clear direct question or imperative.

[verbosity / concise]
  rewrite_instruction: Rewrite as tersely as possible: put the ask first, in one short sentence, with no extra context.

[emotional_charge / calm]
  rewrite_instruction: Rewrite with calm, matter-of-fact, neutral affect and no emotional language.

[native_likeness / native]
  rewrite_instruction: Rewrite in fluent, idiomatic native-English phrasing: natural word order, correct articles and prepositions, and natural collocations. Keep the exact same ask and add no new facts.
\end{lstlisting}

\subsection{Gate A: Intent Preservation}
\label{app:gate-intent}
\begin{lstlisting}[breaklines=true]
You compare two versions of a question asked about an email.
Decide whether the VARIANT asks for EXACTLY the same information as the ORIGINAL — i.e. any answer that correctly answers one would correctly answer the other.
Ignore differences in tone, politeness, directness, length, or wording; judge ONLY the information requested.
Respond with exactly one line: 'Verdict: yes' (same information) or 'Verdict: no' (different information).
\end{lstlisting}

\subsection{Gate B: Style Realization}
\label{app:gate-style}
\begin{lstlisting}[breaklines=true]
You check whether a rewritten question exhibits a specified communication style.
Judge only the style, not the correctness or meaning of the question.
Respond with exactly one line: 'Verdict: yes' (clearly exhibits the style) or 'Verdict: no' (does not).
\end{lstlisting}

\subsection{Fidelity Criteria for All Ten Poles}
\label{app:pole-defs}
Supplied to Gate B as the definition of the target pole. \texttt{native\_likeness} is the grammaticality axis, and \texttt{native} and \texttt{non\_native} are its two poles.
\begin{lstlisting}[breaklines=true]
[directness / direct]
  fidelity_criterion: The request is explicit and on-record — a reader immediately sees exactly what is being asked, as a direct question or command.

[directness / indirect]
  fidelity_criterion: The request is indirect — hedged, hinted, or phrased as a wondering or a stated need — rather than a blunt, bare, on-record question or command. Conventionally-indirect forms ('I was wondering what…', 'I'm trying to figure out…', 'Do you happen to know…') and off-record hints both count; a plain direct question ('What is X?') or an imperative does NOT.

[formality / formal]
  fidelity_criterion: The text reads as a formal/professional register — full words (no contractions or slang) with standard capitalisation and punctuation.

[formality / casual]
  fidelity_criterion: The text reads as casual / chat-like / informal (contractions, slang, or relaxed punctuation), while the underlying ask is unchanged.

[verbosity / concise]
  fidelity_criterion: The text is short and the ask is stated up front.

[verbosity / verbose]
  fidelity_criterion: The text is long and the ask is buried in the middle rather than stated up front, and no new request or factual detail has been added.

[emotional_charge / calm]
  fidelity_criterion: The tone is calm and neutral, with no strong emotional language.

[emotional_charge / charged]
  fidelity_criterion: The text clearly conveys frustration, anxiety, stress, or urgency (not just a plain, neutral question), while the underlying ask is unchanged.

[native_likeness / native]
  fidelity_criterion: The text reads as fluent, idiomatic native English — natural grammar, articles, prepositions, and word choice, with no learner-like errors.

[native_likeness / non_native]
  fidelity_criterion: The text reads as non-native / second-language (L2) English rather than fluent idiomatic native English. It may carry one or more learner-like errors (article, preposition, agreement, verb-tense, or word-order slips) or simply read as awkward or unidiomatic; it need not be heavily broken to count.
\end{lstlisting}

\subsection{Gate C: Human-Likeness}
\label{app:gate-human}
\begin{lstlisting}[breaklines=true]
You judge whether a short message reads as if a REAL PERSON genuinely wrote it, or reads as ARTIFICIAL — machine-generated, templated, or performative.
The message is a QUESTION about an email, rephrased in a particular style. It may legitimately be LONG and DETAILED and name specific people, dates, addresses, or figures — length, detail, and formality are NOT signs of artificiality, so never penalize a message just for being long or thorough.
It may be in any register (formal, casual, emotional, or non-native / learner English) and still be human: judge authenticity and naturalness, NOT fluency, correctness, or politeness. A non-native writer's careful, mostly-fluent question carrying a few uneven slips (missing articles, agreement, tense, or word order) IS human — it need not be heavily or evenly broken.
Mark it artificial ('no') only if it reads as machine-produced: it stacks unnatural intensifiers, uses robotic or templated phrasing or mechanical eye-dialect, or bolts a foreign greeting, honorific, or tag onto otherwise-perfect native English as decoration rather than a genuine change in how the sentence is built.
Respond with exactly one line: 'Verdict: yes' (reads as genuinely human) or 'Verdict: no' (reads as artificial / AI-like).
\end{lstlisting}

\subsection{Verbalized-Sampling User Template}
\label{app:vs-template}
\begin{lstlisting}[breaklines=true]
Produce {k} DIFFERENT rewrites of the question below, each a different {noun} drawn from this list:
{menu}

Make the {k} rewrites as different from one another as possible (different openings and sentence shapes) while each keeps the SAME single answer and every specific detail. Each must follow all the rules in the system message.

Write only the rewritten question itself for each -- never prefix a rewrite with its {noun} name, number, or any label.

Original question:
{question}

Return ONLY a JSON array of exactly {k} strings, e.g. ["...", "..."]. No other text.
\end{lstlisting}

\section{Validation Gates}
\label{app:gates}

The candidate rewrite passes through a three-layer gate before it is admitted. Overall gate-passing rate per dataset is in Table~\ref{tab:gate-yield-total} and per-pole gate-passing rate in Table~\ref{tab:gate-yield-pole}.

\paragraph{Layer 1: deterministic checks.} A candidate is rejected if it is empty, if it is byte-identical to the original after normalization, or if any numeric token present in the original is absent from the variant. Missing capitalized tokens are recorded as a soft signal but do not reject, since names may legitimately be rephrased.

\paragraph{Layer 2: rubric judges.} Gate A (Appendix \ref{app:gate-intent}) asks whether the variant requests exactly the same information. Gate B (Appendix \ref{app:gate-style}) asks whether the variant realizes the target pole, and is given that pole's fidelity criterion from Appendix \ref{app:pole-defs}. Gate C (Appendix \ref{app:gate-human}) asks whether the text reads as something a person would genuinely type. All three answer on a single line so the verdict can be parsed deterministically.

\paragraph{Gate C exceptions.} Gate C is not applied to the verbose pole, where a long message wrapped around a one-line question is inherently machine-like and shortening it would remove the property under test; that pole is admitted on Gates A and B alone. On the non-grammatical pole Gate C is computed and recorded but does not filter, because blocking on it would bias selection toward the mildest, least restructured rewrites.

\paragraph{One low-gate-passing cell.} Yield is above 80\% for 29 of the 30 dataset-pole cells. WorkBench \texttt{indirect} admits 64 of 90, with the loss concentrated at Gate A: WorkBench instructions name a specific record to modify, and an indirect rewrite that hides the operation often also loses the record identifier, at which point the variant is no longer holding the same question intent as the original query. 

\begin{table}[H]
\centering
\small
\setlength{\tabcolsep}{4pt}
\begin{tabular}{lrrr}
\toprule
\textbf{Dataset} & \textbf{Slots} & \textbf{Admitted} & \textbf{\%} \\
\midrule
EnronQA & 4000 & 3943 & 98.6 \\
ToolTalk & 1720 & 1599 & 93.0 \\
WorkBench & 900 & 855 & 95.0 \\
\bottomrule
\end{tabular}
\caption{Overall gate-passing rate per dataset. A slot is one (item, pole) pair; on ToolTalk a slot is one (user turn, pole) pair.}
\label{tab:gate-yield-total}
\end{table}

\begin{table}[H]
\centering
\scriptsize
\setlength{\tabcolsep}{3.5pt}
\begin{tabular}{lllrrrrrrrrr}
\toprule
 & & & \multicolumn{3}{c}{\textbf{EnronQA}} & \multicolumn{3}{c}{\textbf{ToolTalk}} & \multicolumn{3}{c}{\textbf{WorkBench}} \\
\cmidrule(lr){4-6}\cmidrule(lr){7-9}\cmidrule(lr){10-12}
\textbf{Axis} & \textbf{Pole} & \textbf{Role} & \textbf{A} & \textbf{B} & \textbf{C} & \textbf{A} & \textbf{B} & \textbf{C} & \textbf{A} & \textbf{B} & \textbf{C} \\
\midrule
Directness & \texttt{direct} & ref & 99.8 & 100.0 & -- & 93.0 & 91.3 & -- & 100.0 & 100.0 & -- \\
Directness & \texttt{indirect} & con & 100.0 & 100.0 & 100.0 & 83.1 & 94.2 & 99.4 & 73.3 & 94.4 & 98.9 \\
Formality & \texttt{formal} & con & 100.0 & 100.0 & 100.0 & 98.8 & 95.9 & 99.4 & 100.0 & 96.7 & 100.0 \\
Formality & \texttt{casual} & con & 100.0 & 100.0 & 100.0 & 98.8 & 100.0 & 100.0 & 100.0 & 100.0 & 100.0 \\
Verbosity & \texttt{concise} & ref & 99.8 & 90.5 & -- & 93.6 & 92.4 & -- & 100.0 & 97.8 & -- \\
Verbosity & \texttt{verbose} & con & 100.0 & 100.0 & -- & 100.0 & 96.5 & -- & 100.0 & 100.0 & -- \\
Emotion & \texttt{calm} & ref & 99.5 & 100.0 & -- & 99.4 & 97.7 & -- & 97.8 & 98.9 & -- \\
Emotion & \texttt{charged} & con & 100.0 & 98.5 & 100.0 & 96.5 & 99.4 & 100.0 & 100.0 & 100.0 & 100.0 \\
Gramm. & \texttt{grammatical} & ref & 99.8 & 98.8 & -- & 99.4 & 94.8 & -- & 100.0 & 95.6 & -- \\
Gramm. & \texttt{non-grammatical} & con & 100.0 & 100.0 & 76.2 & 100.0 & 100.0 & 80.8 & 100.0 & 100.0 & 94.4 \\
\bottomrule
\end{tabular}
\caption{Per-pole gate-passing rates, in percent of slots. \textbf{A} is intent preservation, \textbf{B} is style realization, and \textbf{C} is human-likeness. Slots per pole are 400 (EnronQA), 50 (ToolTalk), 90 (WorkBench). \emph{Role} is reference (ref) or contrastive (con). Gate C is evaluated only on five contrastive poles: it is not applied to the reference poles or to \texttt{verbose}, shown as \texttt{--}, and on \texttt{non-grammatical} it is recorded without filtering, so that figure describes admitted variants rather than a pass rate. See Appendix~\ref{app:gates}.}
\label{tab:gate-yield-pole}
\end{table}

\section{EnronQA Error Labeling}
\label{app:errors}

The error analysis of \S\ref{subsubsec:stage-rag} covers the headline condition (\texttt{gpt-5.6-terra}, SPLADE, judged by \texttt{gpt-5.4}) for baseline and ten communication style variants poles. There are $118$ generation errors are slots in which the gold email was retrieved and the answer was still judged wrong, so the reader had the evidence in front of it. There are $237$ retrieval failures that are slots in which the gold email never reached the reader. 

\paragraph{Taxonomy.} Each generation error carries exactly one code, assigned under a fixed precedence so that no slot is double-counted. \textbf{E6}, not a genuine reader failure, fires first and covers defective gold answers and judge strictness. \textbf{E3} is an answer faithful to one of the other four retrieved emails rather than the gold one. \textbf{E2b} is a real proposition attached to the wrong role, subject or event, and \textbf{E2a} a field filled with a wrong value. These two combines to \textbf{E2}, unsupported content. \textbf{E4} is an answer of the wrong semantic type, for instance a name where a time was asked for. \textbf{E1} is an answer that is not false but incomplete. \textbf{E5}, false abstention, is a claim that the evidence does not contain the answer when it does. Retrieval failures use three codes: \textbf{R1} reports the asked fact absent, \textbf{R2} hedges, and \textbf{R3} answers confidently anyway.

\paragraph{Procedure.} Labels were assigned by \texttt{claude-opus-5}. For each slot it saw the original question, the rewrite, the gold answer with its accepted alternates, the model answer, the baseline answer, and the full text of every retrieved email, and returned one code with a written rationale. One author then reviewed all $118$ labels against the same evidence and recoded $23$ where the rationale did not support the code under the precedence rule.

\paragraph{Slices.} An error is \emph{style-induced} when the unperturbed request answered the item correctly and the rewrite broke it ($44$ slots), \emph{persistent} when the baseline was already wrong ($65$), and \emph{baseline} when it is the unperturbed run's own error ($9$). Table~\ref{app:error-dist} sets the style-induced slice against the other two pooled, which isolates the kind of failure a rewrite produces. Two codes appear only in the style-induced slice: answering in the wrong semantic type ($4$ slots) and falsely reporting the answer absent ($2$), both of which misread the request. Answering from another retrieved email runs the other way, $9.1\%$ of the style-induced slice against $27.0\%$ of the rest. Unsupported content is the largest category in both, $43.2\%$ and $47.3\%$.

\paragraph{Missing evidence.} The $237$ retrieval failures are coded separately. The reader reports the fact absent in $55$ of them ($23.2\%$), hedges in $6$ ($2.5\%$), and answers confidently anyway in $176$ ($74.3\%$). It behaves the same way whether the item was never retrievable under the unperturbed request or the rewrite pushed the gold email out of the top five.

\section{Estimators}
\label{app:stats}

Notation follows \S\ref{subsec:stats}: $d_i$ is the paired difference for item $i$ and $n$ the number of items present in both the base and variant conditions.

\paragraph{Paired bootstrap.} We test $\bar{d} = \frac{1}{n}\sum_i d_i$ by resampling item indices with replacement. For $b = 1,\dots,B$ with $B = 10{,}000$ and a fixed seed,
\begin{equation}
\bar{d}^{*(b)} \;=\; \frac{1}{n}\sum_{j=1}^{n} d_{\,i_j^{(b)}}, \qquad i_j^{(b)} \sim \mathrm{Unif}\{1,\dots,n\}.
\label{eq:boot}
\end{equation}
The 95\% interval is the $2.5$ and $97.5$ percentiles of $\{\bar{d}^{*(b)}\}$, and the two-sided $p$-value comes from the proportion of resampled means on either side of zero, capped at $1$:
\begin{equation}
\begin{aligned}
\hat{F}_{\lessgtr 0} &= \tfrac{1}{B}\textstyle\sum_b \mathbb{1}\big[\bar{d}^{*(b)} \lessgtr 0\big], \\[2pt]
p &= 2\min\!\big(\hat{F}_{\le 0},\; \hat{F}_{\ge 0}\big).
\end{aligned}
\label{eq:bootp}
\end{equation}

\paragraph{Exact McNemar test.} On EnronQA each item yields one binary outcome from a single run, so the paired difference is $-1$, $0$ or $+1$ and only the discordant pairs carry information. With $b$ items correct at baseline but wrong under the variant, and $c$ the reverse,
\begin{equation}
p \;=\; 2\!\!\sum_{i=0}^{\min(b,c)} \binom{b+c}{i} \left(\tfrac{1}{2}\right)^{b+c}.
\label{eq:mcnemar}
\end{equation}
This conditions on $b+c$ and is exact at any sample size, which matters because $b+c$ is often small: over the EnronQA grid it has a median of $18$ and falls below $10$ in $20$ of $114$ cells. A bootstrap over all $n$ items would treat a handful of same-signed events as a stable mean and is correspondingly anti-conservative in that regime.

\section{Compute Resources} 

Experiments were conducted using API-accessed language models and standard retrieval pipelines. Retrieval experiments were run on a local machine with 64 GB RAM. No model training or fine-tuning was performed. All RAG and agentic evaluations consisted exclusively of inference-time API calls. Statistical analyses were performed using standard CPU-based computation.

\begin{table}[H]
\centering
\small
\setlength{\tabcolsep}{4pt}
\begin{tabular}{lccc}
\toprule
\textbf{Benchmark} & terra & mini & Kimi \\
\midrule
EnronQA & $1{\times}3$ & $1{\times}3$ & $1{\times}3$ \\
ToolTalk & $K{=}3$ & $K{=}3$ & $K{=}3$ \\
WorkBench & $K{=}3$ & -- & -- \\
\bottomrule
\end{tabular}
\caption{Measured conditions per benchmark and model. Replicate runs per condition for each benchmark and model. All 14 conditions the design defines are measured in every cell shown: ten communication-style poles and four dialect varieties, each paired against that model's unperturbed baseline. EnronQA is one inference run scored under three retrievers, written $1{\times}3$. The headline model is complete on all three benchmarks. WorkBench was not completed on the other two models and is reported on the headline model alone, so the cross-model comparison in Table~\ref{tab:generality} is scoped to the two benchmarks where all three models were measured.}
\label{app:coverage}
\end{table}

\begin{table}[H]
\centering
\scriptsize
\setlength{\tabcolsep}{3pt}
\begin{tabular}{llrrrrrr}
\toprule
\textbf{Benchmark} & \textbf{Condition} & $n$ & base & cond & $\Delta$ & 95\% CI & $p$ \\
\midrule
\multirow{14}{*}{EnronQA} & \texttt{direct} & 399 & 92.7 & 93.0 & +0.3\sig{} & -- & 1.000 \\
 & \texttt{indirect} & 400 & 92.8 & 89.5 & \textbf{-3.2}\sig{**} & -- & 0.007 \\
 & \texttt{formal} & 400 & 92.8 & 91.8 & -1.0\sig{} & -- & 0.388 \\
 & \texttt{casual} & 400 & 92.8 & 91.8 & -1.0\sig{} & -- & 0.344 \\
 & \texttt{concise} & 358 & 91.9 & 91.9 & +0.0\sig{} & -- & 1.000 \\
 & \texttt{verbose} & 400 & 92.8 & 89.8 & \textbf{-3.0}\sig{*} & -- & 0.012 \\
 & \texttt{calm} & 398 & 92.7 & 92.2 & -0.5\sig{} & -- & 0.625 \\
 & \texttt{charged} & 394 & 93.1 & 93.7 & +0.5\sig{} & -- & 0.754 \\
 & \texttt{grammatical} & 394 & 92.6 & 92.6 & +0.0\sig{} & -- & 1.000 \\
 & \texttt{non-grammatical} & 399 & 93.0 & 91.2 & -1.8\sig{} & -- & 0.167 \\
 & \texttt{Indian} & 381 & 92.7 & 88.7 & \textbf{-3.9}\sig{**} & -- & 0.001 \\
 & \texttt{Singapore} & 382 & 92.7 & 90.1 & \textbf{-2.6}\sig{*} & -- & 0.021 \\
 & \texttt{Black S. African} & 377 & 92.6 & 89.9 & \textbf{-2.7}\sig{*} & -- & 0.021 \\
 & \texttt{Nigerian} & 374 & 92.5 & 89.8 & \textbf{-2.7}\sig{*} & -- & 0.031 \\
\midrule
\multirow{14}{*}{ToolTalk} & \texttt{direct} & 50 & 52.7 & 50.7 & -2.0\sig{} & [-9.3, +4.0] & 0.633 \\
 & \texttt{indirect} & 49 & 52.4 & 29.9 & \textbf{-22.4}\sig{***} & [-35.4, -10.9] & 0.000 \\
 & \texttt{formal} & 48 & 50.7 & 36.8 & \textbf{-13.9}\sig{***} & [-22.9, -5.6] & 0.000 \\
 & \texttt{casual} & 50 & 52.7 & 48.7 & -4.0\sig{} & [-12.7, +4.0] & 0.343 \\
 & \texttt{concise} & 48 & 50.7 & 41.7 & \textbf{-9.0}\sig{**} & [-16.7, -2.1] & 0.010 \\
 & \texttt{verbose} & 50 & 52.7 & 48.0 & -4.7\sig{} & [-13.3, +4.0] & 0.266 \\
 & \texttt{calm} & 50 & 52.7 & 54.0 & +1.3\sig{} & [-4.7, +7.3] & 0.699 \\
 & \texttt{charged} & 50 & 52.7 & 44.0 & -8.7\sig{} & [-18.7, +0.7] & 0.080 \\
 & \texttt{grammatical} & 49 & 51.7 & 46.9 & -4.8\sig{} & [-10.9, +1.4] & 0.141 \\
 & \texttt{non-grammatical} & 50 & 52.7 & 43.3 & \textbf{-9.3}\sig{*} & [-18.7, -0.0] & 0.048 \\
 & \texttt{Indian} & 50 & 52.7 & 38.0 & \textbf{-14.7}\sig{**} & [-25.3, -4.7] & 0.004 \\
 & \texttt{Singapore} & 50 & 52.7 & 44.0 & \textbf{-8.7}\sig{*} & [-18.0, -0.7] & 0.028 \\
 & \texttt{Black S. African} & 50 & 52.7 & 52.7 & +0.0\sig{} & [-7.3, +7.3] & 1.000 \\
 & \texttt{Nigerian} & 49 & 51.7 & 44.9 & -6.8\sig{} & [-14.3, +0.7] & 0.057 \\
\midrule
\multirow{14}{*}{WorkBench} & \texttt{direct} & 90 & 90.7 & 88.1 & -2.6\sig{} & [-8.1, +2.6] & 0.339 \\
 & \texttt{indirect} & 90 & 90.7 & 80.0 & \textbf{-10.7}\sig{**} & [-17.8, -4.1] & 0.002 \\
 & \texttt{formal} & 90 & 90.7 & 78.5 & \textbf{-12.2}\sig{**} & [-21.1, -4.1] & 0.003 \\
 & \texttt{casual} & 90 & 90.7 & 90.4 & -0.4\sig{} & [-5.6, +4.8] & 0.889 \\
 & \texttt{concise} & 90 & 90.7 & 94.8 & +4.1\sig{} & [-0.4, +8.9] & 0.083 \\
 & \texttt{verbose} & 90 & 90.7 & 88.9 & -1.9\sig{} & [-7.0, +3.3] & 0.486 \\
 & \texttt{calm} & 90 & 90.7 & 88.9 & -1.9\sig{} & [-6.7, +3.0] & 0.478 \\
 & \texttt{charged} & 90 & 90.7 & 82.2 & \textbf{-8.5}\sig{*} & [-15.6, -1.9] & 0.012 \\
 & \texttt{grammatical} & 90 & 90.7 & 88.1 & -2.6\sig{} & [-8.5, +2.6] & 0.382 \\
 & \texttt{non-grammatical} & 90 & 90.7 & 86.7 & -4.1\sig{} & [-9.6, +1.5] & 0.155 \\
 & \texttt{Indian} & 90 & 90.7 & 88.9 & -1.9\sig{} & [-7.0, +3.3] & 0.509 \\
 & \texttt{Singapore} & 90 & 90.7 & 83.7 & \textbf{-7.0}\sig{*} & [-14.1, -0.0] & 0.047 \\
 & \texttt{Black S. African} & 90 & 90.7 & 90.0 & -0.7\sig{} & [-6.3, +4.8] & 0.805 \\
 & \texttt{Nigerian} & 90 & 90.7 & 90.4 & -0.4\sig{} & [-5.2, +4.1] & 0.870 \\
\bottomrule
\end{tabular}
\caption{Per-cell detail for all three benchmarks, headline model. Every cell of Table~\ref{tab:main-style} with its paired item count, matched baseline, condition value, interval and raw $p$. \emph{base} is the same model's score on exactly the items that condition covers, so base${+}\Delta{=}$cond on every row, and \emph{base} differs slightly between rows wherever the invariance gate dropped items. All values are percentages on the benchmark's own metric. $p$ is each benchmark's reported test (\S\ref{subsec:stats}): the exact McNemar test on EnronQA and a paired bootstrap over items elsewhere, with the interval from that same bootstrap. The exact test conditions on the discordant pairs and yields no interval on this scale, so the EnronQA rows carry $p$ alone. \sigl{*}$p{<}.05$, \sigl{**}$p{<}.01$, \sigl{***}$p{<}.001$.}
\label{app:full-detail}
\end{table}

\begin{table}[H]
\centering
\small
\setlength{\tabcolsep}{6pt}
\begin{tabular}{llrrrr}
\toprule
\textbf{Code} & \textbf{Failure mode} & \multicolumn{2}{c}{\textbf{style-induced}} & \multicolumn{2}{c}{\textbf{reader}} \\
\cmidrule(lr){3-4}\cmidrule(lr){5-6}
 & & $n$ & \% & $n$ & \% \\
\midrule
\textbf{E1} & incomplete answer & 6 & 13.6 & 7 & 9.5 \\
\textbf{E2} & unsupported content & 19 & 43.2 & 35 & 47.3 \\
\textbf{E3} & wrong source email & 4 & 9.1 & 20 & 27.0 \\
\textbf{E4} & wrong answer mode & 4 & 9.1 & 0 & 0.0 \\
\textbf{E5} & false abstention & 2 & 4.5 & 0 & 0.0 \\
\textbf{E6} & not a reader failure & 9 & 20.5 & 12 & 16.2 \\
\midrule
& \emph{total} & 44 & 100.0 & 74 & 100.0 \\
\bottomrule
\end{tabular}
\caption{EnronQA error profile, rewrite-induced against the reader's own failures. The $118$ labeled generation errors on EnronQA (\texttt{gpt-5.6-terra} $\times$ SPLADE), split by whether the rewrite created the error. \emph{style-induced} errors are those on items the unperturbed request answered correctly; \emph{reader} pools the errors the unperturbed request made itself ($9$) with those it made on items it was already failing ($65$). Code definitions are in Appendix~\ref{app:errors}.}
\label{app:error-dist}
\end{table}

\begin{table}[H]
\centering
\small
\setlength{\tabcolsep}{6pt}
\begin{tabular}{lrrrrr}
\toprule
\textbf{Group} & $n$ & \textbf{Base} & \textbf{ref} & \textbf{con} & \textbf{dia} \\
\midrule
\multicolumn{6}{@{}l}{\emph{by complexity bin}} \\
\addlinespace[1pt]
\quad simple \emph{(control)} & 100 & 93.0 & 0.28 & 1.62 & 3.13 \\
\quad 3-hop & 100 & 90.0 & 1.17 & 4.26 & 5.34 \\
\quad multi-question & 100 & 90.0 & 1.71 & 4.08 & 5.36 \\
\quad threaded & 100 & 98.0 & 0.00 & 1.88 & 2.91 \\
\addlinespace[3pt]
\multicolumn{6}{@{}l}{\emph{by reasoning type}} \\
\addlinespace[1pt]
\quad extraction & 126 & 92.1 & 0.89 & 4.74 & 5.75 \\
\quad bridging & 74 & 95.9 & 0.73 & 1.88 & 2.99 \\
\quad causal & 61 & 98.4 & 0.87 & 1.67 & 4.64 \\
\quad temporal & 57 & 94.7 & 1.43 & 2.17 & 3.02 \\
\quad aggregation & 45 & 80.0 & 0.00 & 0.00 & 1.40 \\
\quad comparative & 37 & 91.9 & 0.00 & 5.42 & 5.00 \\
\bottomrule
\end{tabular}
\caption{EnronQA fragility by question complexity and reasoning type. Headline condition (\texttt{gpt-5.6-terra}, SPLADE). \textbf{Base} is unperturbed accuracy. The three breakage columns are the percentage of (item, pole) slots that the baseline answered correctly and the rewrite broke, over the four reference poles (\textbf{ref}), the six contrastive style poles (\textbf{con}) and the four dialect varieties (\textbf{dia}).}
\label{app:complexity}
\end{table}

\begin{figure}[H]
\centering
\includegraphics[width=\columnwidth]{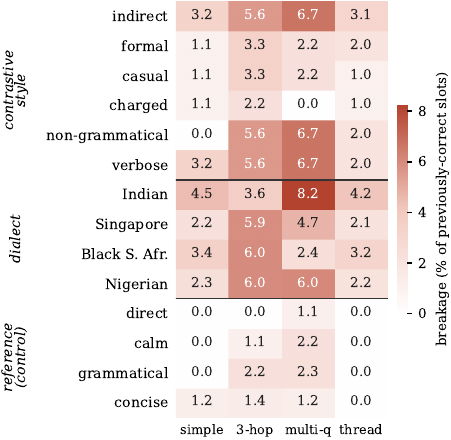}
\caption{EnronQA breakage by pole and question-complexity bin. Columns are the four sampling bins: \emph{simple} (the
control), \emph{3-hop}, \emph{multi-question} and \emph{threaded}. Each cell is the
percentage of slots in that bin which the unperturbed request answered correctly and the
rewrite broke.}
\label{app:fig-complexity}
\end{figure}

\begin{table}[H]
\centering
\scriptsize
\setlength{\tabcolsep}{1.5pt}
\begin{tabular}{llrrrrrrrrr}
\toprule
& & \multicolumn{3}{c}{terra} & \multicolumn{3}{c}{mini} & \multicolumn{3}{c}{Kimi} \\
\cmidrule(lr){3-5}\cmidrule(lr){6-8}\cmidrule(lr){9-11}
\textbf{Pole} & & B & G & S & B & G & S & B & G & S \\
\midrule
\texttt{direct} & ref & -0.8\sig{} & +0.3\sig{} & +0.3\sig{} & +0.8\sig{} & -0.3\sig{} & -0.8\sig{} & -0.8\sig{} & +0.0\sig{} & -0.3\sig{} \\
\texttt{indirect} & con & -2.3\sig{} & \textbf{-3.8}\sig{**} & \textbf{-3.2}\sig{**} & \textbf{-3.5}\sig{*} & \textbf{-4.8}\sig{***} & \textbf{-3.2}\sig{*} & \textbf{-3.8}\sig{**} & \textbf{-3.0}\sig{*} & -1.5\sig{} \\
\texttt{calm} & ref & +0.3\sig{} & -0.3\sig{} & -0.5\sig{} & +0.5\sig{} & -1.0\sig{} & -0.3\sig{} & -1.0\sig{} & -0.3\sig{} & +0.3\sig{} \\
\texttt{charged} & con & -2.0\sig{} & \textbf{-2.8}\sig{*} & +0.5\sig{} & -1.0\sig{} & -2.5\sig{} & -1.0\sig{} & \textbf{-3.3}\sig{**} & -1.5\sig{} & -0.3\sig{} \\
\texttt{casual} & con & -1.0\sig{} & \textbf{-3.2}\sig{*} & -1.0\sig{} & -0.8\sig{} & -2.8\sig{} & -1.5\sig{} & \textbf{-3.0}\sig{**} & \textbf{-3.3}\sig{*} & -1.2\sig{} \\
\texttt{formal} & con & -1.0\sig{} & -1.2\sig{} & -1.0\sig{} & -0.8\sig{} & -1.5\sig{} & -1.0\sig{} & -0.8\sig{} & -1.2\sig{} & +0.5\sig{} \\
\texttt{grammatical} & ref & +1.0\sig{} & +0.8\sig{} & +0.0\sig{} & +0.0\sig{} & +1.0\sig{} & -0.5\sig{} & -1.0\sig{} & +0.5\sig{} & +1.8\sig{} \\
\texttt{non-grammatical} & con & -2.5\sig{} & -2.5\sig{} & -1.8\sig{} & -1.8\sig{} & \textbf{-3.2}\sig{**} & -2.5\sig{} & \textbf{-2.8}\sig{*} & -1.5\sig{} & +1.0\sig{} \\
\texttt{concise} & ref & -0.3\sig{} & -1.1\sig{} & +0.0\sig{} & +0.6\sig{} & -0.3\sig{} & -0.6\sig{} & -1.7\sig{} & -0.8\sig{} & +0.3\sig{} \\
\texttt{verbose} & con & \textbf{-26.0}\sig{***} & \textbf{-4.5}\sig{***} & \textbf{-3.0}\sig{*} & \textbf{-25.0}\sig{***} & \textbf{-5.0}\sig{***} & -1.5\sig{} & \textbf{-27.1}\sig{***} & \textbf{-3.3}\sig{*} & -0.5\sig{} \\
\bottomrule
\end{tabular}
\caption{EnronQA accuracy $\Delta$ for every answerer $\times$ retriever condition. Each answerer has three columns, one per retriever: B = BM25, G = BGE-M3, S = SPLADE. Cells are accuracy $\Delta$ against that condition's own unperturbed baseline. \sigl{*}$p{<}.05$, \sigl{**}$p{<}.01$, \sigl{***}$p{<}.001$.}
\label{app:grid}
\end{table}

\begin{table}[H]
\centering
\small
\setlength{\tabcolsep}{4pt}
\begin{tabular}{llrrrr}
\toprule
\textbf{Model} & \textbf{Retr.} & R@5 & MRR & \multicolumn{2}{c}{\textbf{poles with $p<.05$}} \\
\cmidrule(lr){5-6}
 & & & & R@5 & acc \\
\midrule
terra & BM25 & 92.8 & 0.841 & 1/6 & 1/6 \\
terra & BGE-M3 & 88.8 & 0.795 & 3/6 & 4/6 \\
terra$\star$ & SPLADE & 93.5 & 0.854 & 1/6 & 2/6 \\
mini & BM25 & 92.8 & 0.841 & 1/6 & 2/6 \\
mini & BGE-M3 & 88.8 & 0.795 & 3/6 & 3/6 \\
mini & SPLADE & 93.5 & 0.854 & 1/6 & 1/6 \\
Kimi & BM25 & 92.8 & 0.841 & 1/6 & 5/6 \\
Kimi & BGE-M3 & 88.8 & 0.795 & 3/6 & 3/6 \\
Kimi & SPLADE & 93.5 & 0.854 & 1/6 & 0/6 \\
\bottomrule
\end{tabular}
\caption{Retrieval quality and style-invariance of retrieval, by condition. $\star$ marks the headline condition. R@5 and MRR are the retrieval quality of the unperturbed queries under that retriever. The two counts are the number of the six contrastive poles reaching $p<.05$ on gold-email recall and on answer accuracy.}
\label{app:equiv}
\end{table}

\begin{table}[H]
\centering
\small
\setlength{\tabcolsep}{3pt}
\begin{tabular}{lrrrrrr}
\toprule
& \multicolumn{3}{c}{\textbf{EnronQA}} & \multicolumn{3}{c}{\textbf{ToolTalk}} \\
\cmidrule(lr){2-4}\cmidrule(lr){5-7}
\textbf{Variety} & terra & mini & Kimi & terra & mini & Kimi \\
\midrule
Indian & \textbf{-3.9}\sig{**} & \textbf{-3.1}\sig{**} & -1.8\sig{} & \textbf{-14.7}\sig{**} & \textbf{-16.0}\sig{*} & -6.8\sig{} \\
Singapore & \textbf{-2.6}\sig{*} & \textbf{-6.0}\sig{***} & -1.8\sig{} & \textbf{-8.7}\sig{*} & -10.7\sig{} & -0.7\sig{} \\
Nigerian & \textbf{-2.7}\sig{*} & -2.7\sig{} & -1.3\sig{} & -6.8\sig{} & -6.8\sig{} & +2.7\sig{} \\
Black S. Afr. & \textbf{-2.7}\sig{*} & -2.1\sig{} & -0.5\sig{} & +0.0\sig{} & \textbf{-18.0}\sig{**} & -3.3\sig{} \\
\bottomrule
\end{tabular}
\caption{Dialect $\Delta$ for four attested varieties, three answerer models, two benchmarks. Each cell is the paired $\Delta$ on that benchmark's primary metric (EnronQA accuracy, ToolTalk conversation success) against the same model's own unperturbed baseline. EnronQA is SPLADE for every model. Restricted to the two benchmarks on which all three models were run; the headline model's dialect results on all three are in Table~\ref{tab:main-style}. Varieties are generated by the rule-based transducer at density~1.0 and are analysed separately, not pooled. Each benchmark uses the test matched to its outcome (\S\ref{subsec:stats}): the exact McNemar test on EnronQA, a paired bootstrap over items with $B{=}10{,}000$ elsewhere. Raw $p$, no multiplicity correction. \sigl{*}$p{<}.05$, \sigl{**}$p{<}.01$, \sigl{***}$p{<}.001$.}
\label{app:dialect-models}
\end{table}

\begin{table}[H]
\centering
\small
\setlength{\tabcolsep}{4pt}
\begin{tabular}{llrrr}
\toprule
\textbf{Pole} & & \textbf{EnronQA} & \textbf{ToolTalk} & \textbf{WorkBench} \\
\midrule
\texttt{direct} & ref & 65.4 & 64.6 & 31.1 \\
\texttt{calm} & ref & 70.1 & 72.5 & 31.0 \\
\texttt{grammatical} & ref & 60.4 & 76.4 & 25.9 \\
\texttt{concise} & ref & 61.7 & 61.5 & 27.6 \\
\addlinespace[2pt]
\texttt{indirect} & con & \textbf{12.5} & \textbf{5.0} & \textbf{4.7} \\
\texttt{formal} & con & 16.2 & 27.4 & 15.5 \\
\texttt{casual} & con & 24.0 & 54.7 & 19.3 \\
\texttt{charged} & con & 36.1 & 55.8 & 22.2 \\
\texttt{non-grammatical} & con & 38.0 & 48.3 & 25.6 \\
\texttt{verbose} & con & 17.0 & 35.5 & 14.4 \\
\bottomrule
\end{tabular}
\caption{Retention of the original head verb as the main predicate, per pole and benchmark. Each cell is the percentage of that pole's variants in which the head verb of the original request is still the main predicate, matched on lemma and read from a dependency parse. The complement is split between deletion, demotion to a subordinate clause, and survival in another role. Rows are the four reference poles followed by the six contrastive poles; \emph{ref} and \emph{con} mark the role. Bold marks the lowest cell in each column. Levels are not comparable between columns: WorkBench instructions are terse imperatives, so even a reference rewrite often replaces the verb, which lowers the whole column. Computed over all gate-passing slots (EnronQA $n{=}$3941), (ToolTalk $n{=}$1599), (WorkBench $n{=}$855); ToolTalk is measured per turn because a main clause is a property of one message.}
\label{app:verb-fate}
\end{table}

\begin{table}[H]
\centering
\small
\setlength{\tabcolsep}{6pt}
\begin{tabular}{lrrr}
\toprule
\textbf{Quartile} & \textbf{Length ratio} & $n$ & \textbf{R@5} \\
\midrule
Q1 & 1.63--2.43$\times$ & 94 & 95.7 \\
Q2 & 2.44--2.88$\times$ & 91 & 82.4 \\
Q3 & 2.88--3.55$\times$ & 93 & 57.0 \\
Q4 & 3.56--16.51$\times$ & 92 & 44.6 \\
\bottomrule
\end{tabular}
\caption{Gold-email recall under \texttt{verbose}, by how much padding the rewrite added. Slots are the \texttt{verbose} variants of EnronQA questions whose gold email was already in the top~5 for the unmodified request under BM25 ($n{=}370$ of 400), split into quartiles by how many times longer the rewrite is than its own original.}
\label{app:verbose-gradient}
\end{table}

\begin{table}[H]
\centering
\small
\setlength{\tabcolsep}{2.5pt}
\begin{tabular}{llrr}
\toprule
\textbf{Pole} & & \textbf{r@5} & \textbf{mrr} \\
\midrule
\texttt{direct} & ref & -0.25\sig{} & -0.01\sig{} \\
\texttt{indirect} & con & \textbf{-1.75}\sig{*} & \textbf{-0.03}\sig{**} \\
\texttt{calm} & ref & -0.25\sig{} & -0.00\sig{} \\
\texttt{charged} & con & +0.00\sig{} & -0.01\sig{} \\
\texttt{casual} & con & -0.75\sig{} & -0.01\sig{} \\
\texttt{formal} & con & -0.25\sig{} & -0.01\sig{} \\
\texttt{grammatical} & ref & +0.25\sig{} & -0.00\sig{} \\
\texttt{non-grammatical} & con & -0.25\sig{} & -0.01\sig{} \\
\texttt{concise} & ref & -0.28\sig{} & \textbf{-0.01}\sig{***} \\
\texttt{verbose} & con & -0.25\sig{} & \textbf{-0.03}\sig{***} \\
\bottomrule
\end{tabular}
\caption{EnronQA intermediate signals, headline condition. \textbf{r@5} is gold-email recall at $k{=}5$ and \textbf{mrr} its mean reciprocal rank. Paired $\Delta$ against the unperturbed request. \sigl{*}$p{<}.05$, \sigl{**}$p{<}.01$, \sigl{***}$p{<}.001$.}
\label{app:mech-enronqa}
\end{table}

\begin{table}[H]
\centering
\small
\setlength{\tabcolsep}{2.5pt}
\begin{tabular}{llrr}
\toprule
\textbf{Pole} & & \textbf{recall} & \textbf{bad\_action\_rate} \\
\midrule
\texttt{direct} & ref & -2.34\sig{} & +2.54\sig{} \\
\texttt{indirect} & con & \textbf{-11.28}\sig{***} & +2.96\sig{} \\
\texttt{calm} & ref & +0.53\sig{} & -1.70\sig{} \\
\texttt{charged} & con & -2.94\sig{} & +3.71\sig{} \\
\texttt{casual} & con & -2.63\sig{} & +0.71\sig{} \\
\texttt{formal} & con & \textbf{-10.32}\sig{***} & -1.94\sig{} \\
\texttt{grammatical} & ref & -0.06\sig{} & \textbf{-4.16}\sig{*} \\
\texttt{non-grammatical} & con & \textbf{-5.91}\sig{**} & +0.29\sig{} \\
\texttt{concise} & ref & -2.76\sig{} & +0.01\sig{} \\
\texttt{verbose} & con & -2.45\sig{} & +0.24\sig{} \\
\bottomrule
\end{tabular}
\caption{ToolTalk intermediate signals, headline condition. \textbf{recall} is the fraction of required ground-truth API calls the agent made and \textbf{bad\_action\_rate} the rate of calls with no ground-truth counterpart. \sigl{**}$p{<}.01$, \sigl{***}$p{<}.001$.}
\label{app:mech-tooltalk}
\end{table}

\begin{table}[H]
\centering
\small
\setlength{\tabcolsep}{2.5pt}
\begin{tabular}{llrrr}
\toprule
\textbf{Pole} & & \textbf{callF1} & \textbf{noCall} & \textbf{sideEffect} \\
\midrule
\texttt{direct} & ref & -1.23\sig{} & +0.37\sig{} & +1.85\sig{} \\
\texttt{indirect} & con & \textbf{-8.42}\sig{***} & +1.11\sig{} & -2.59\sig{} \\
\texttt{calm} & ref & -1.49\sig{} & +0.37\sig{} & +1.11\sig{} \\
\texttt{charged} & con & \textbf{-7.45}\sig{***} & +1.11\sig{} & -2.22\sig{} \\
\texttt{casual} & con & \textbf{-14.75}\sig{***} & +0.00\sig{} & -0.37\sig{} \\
\texttt{formal} & con & \textbf{-10.46}\sig{***} & \textbf{+11.85}\sig{***} & \textbf{-4.07}\sig{*} \\
\texttt{grammatical} & ref & \textbf{-3.67}\sig{*} & +0.37\sig{} & -0.00\sig{} \\
\texttt{non-grammatical} & con & \textbf{-4.47}\sig{***} & +1.85\sig{} & -1.11\sig{} \\
\texttt{concise} & ref & +0.83\sig{} & -0.37\sig{} & -3.70\sig{} \\
\texttt{verbose} & con & -0.99\sig{} & -0.37\sig{} & +1.85\sig{} \\
\bottomrule
\end{tabular}
\caption{WorkBench intermediate signals, headline condition. \textbf{callF1} is the F1 between the agent's predicted tool calls and the ground-truth ones, \textbf{noCall} the rate at which no tool call is issued at all, and \textbf{sideEffect} the rate at which execution leaves an unwanted change to the sandbox. \sigl{*}$p{<}.05$, \sigl{**}$p{<}.01$, \sigl{***}$p{<}.001$.}
\label{app:mech-workbench}
\end{table}

\begin{figure}[H]
\centering
\includegraphics[width=\textwidth]{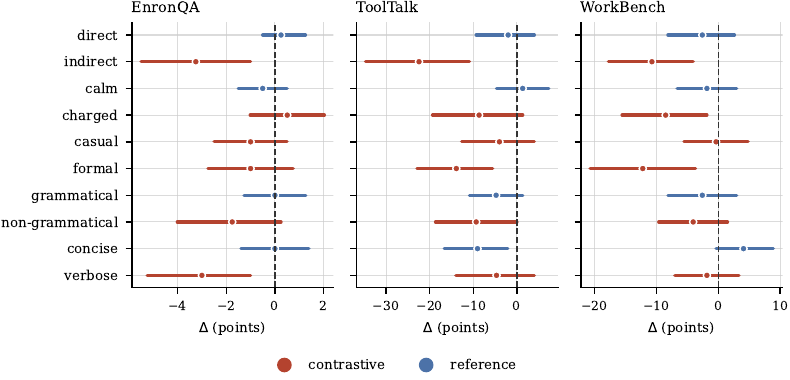}
\caption{All ten poles in the headline condition of each system, with 95\% paired bootstrap confidence intervals. Reference poles are the four meaning-preserving rewrites, contrastive poles the six styled ones. Metrics differ across panels and are not commensurable.}
\label{app:forest}
\end{figure}

\FloatBarrier
\clearpage
\section*{NeurIPS Paper Checklist}

%%% BEGIN INSTRUCTIONS %%%
The checklist is designed to encourage best practices for responsible machine learning research, addressing issues of reproducibility, transparency, research ethics, and societal impact. Do not remove the checklist: {\bf The papers not including the checklist will be desk rejected.} The checklist should follow the references and follow the (optional) supplemental material.  The checklist does NOT count towards the page
limit. 

Please read the checklist guidelines carefully for information on how to answer these questions. For each question in the checklist:
\begin{itemize}
    \item You should answer \answerYes{}, \answerNo{}, or \answerNA{}.
    \item \answerNA{} means either that the question is Not Applicable for that particular paper or the relevant information is Not Available.
    \item Please provide a short (1--2 sentence) justification right after your answer (even for \answerNA). 
   % \item {\bf The papers not including the checklist will be desk rejected.}
\end{itemize}

{\bf The checklist answers are an integral part of your paper submission.} They are visible to the reviewers, area chairs, senior area chairs, and ethics reviewers. You will also be asked to include it (after eventual revisions) with the final version of your paper, and its final version will be published with the paper.

The reviewers of your paper will be asked to use the checklist as one of the factors in their evaluation. While \answerYes{} is generally preferable to \answerNo{}, it is perfectly acceptable to answer \answerNo{} provided a proper justification is given (e.g., error bars are not reported because it would be too computationally expensive'' or ``we were unable to find the license for the dataset we used''). In general, answering \answerNo{} or \answerNA{} is not grounds for rejection. While the questions are phrased in a binary way, we acknowledge that the true answer is often more nuanced, so please just use your best judgment and write a justification to elaborate. All supporting evidence can appear either in the main paper or the supplemental material, provided in appendix. If you answer \answerYes{} to a question, in the justification please point to the section(s) where related material for the question can be found.

IMPORTANT, please:
\begin{itemize}
    \item {\bf Delete this instruction block, but keep the section heading ``NeurIPS Paper Checklist"},
    \item  {\bf Keep the checklist subsection headings, questions/answers and guidelines below.}
    \item {\bf Do not modify the questions and only use the provided macros for your answers}.
\end{itemize}

%%% END INSTRUCTIONS %%%

\begin{enumerate}

\item {\bf Claims}
    \item[] Question: Do the main claims made in the abstract and introduction accurately reflect the paper's contributions and scope?
    \item[] Answer: \answerYes{} % Replace by \answerYes{}, \answerNo{}, or \answerNA{}.
    \item[] Justification: The abstract and introduction state the paper's main claims: that communication-style and dialect variation can reduce the reliability of email assistants, that indirect requests are the most consistently harmful condition, and that agentic failures are primarily omissions of required actions. These claims are supported by the experimental results.

\item {\bf Limitations}
    \item[] Question: Does the paper discuss the limitations of the work performed by the authors?
    \item[] Answer: \answerYes{} % Replace by \answerYes{}, \answerNo{}, or \answerNA{}.
    \item[] Justification: The paper includes a Limitations section (Section 6) that discusses dialect generation at a single feature density, incomplete benchmark coverage, imperfect reference controls, compute constraints, and the absence of mitigation experiments.

\item {\bf Theory assumptions and proofs}
    \item[] Question: For each theoretical result, does the paper provide the full set of assumptions and a complete (and correct) proof?
    \item[] Answer: \answerNA{} % Replace by \answerYes{}, \answerNo{}, or \answerNA{}.
    \item[] Justification: The paper does not present theoretical results, theorems, or proofs. Its contributions are methodological and analytical.

    \item {\bf Experimental result reproducibility}
    \item[] Question: Does the paper fully disclose all the information needed to reproduce the main experimental results of the paper to the extent that it affects the main claims and/or conclusions of the paper (regardless of whether the code and data are provided or not)?
    \item[] Answer: \answerYes{} % Replace by \answerYes{}, \answerNo{}, or \answerNA{}.
    \item[] Justification: The paper describes the datasets, variant-generation procedures, validation gates, evaluation protocols, hyperparameters, metrics, and statistical tests needed to reproduce the experiments. Additional implementation details, prompts, and validation procedures are provided in the appendix.

\item {\bf Open access to data and code}
    \item[] Question: Does the paper provide open access to the data and code, with sufficient instructions to faithfully reproduce the main experimental results, as described in supplemental material?
    \item[] Answer: \answerNo{} % Replace by \answerYes{}, \answerNo{}, or \answerNA{}.
    \item[] Justification: The experiments use a combination of public benchmarks and proprietary model APIs. While the paper provides detailed methodological descriptions, anonymized code and artifacts are not included with the current submission.

\item {\bf Experimental setting/details}
    \item[] Question: Does the paper specify all the training and test details (e.g., data splits, hyperparameters, how they were chosen, type of optimizer) necessary to understand the results?
    \item[] Answer: \answerYes{} % Replace by \answerYes{}, \answerNo{}, or \answerNA{}.
    \item[] Justification: The paper describes benchmark selection, sampling procedures, models, retrievers, prompting methods, validation criteria, evaluation metrics, and statistical testing procedures. Implementation details are provided in the Experiment section and in the appendix.

\item {\bf Experiment statistical significance}
    \item[] Question: Does the paper report error bars suitably and correctly defined or other appropriate information about the statistical significance of the experiments?
    \item[] Answer: \answerYes{} % Replace by \answerYes{}, \answerNo{}, or \answerNA{}.
    \item[] Justification: The paper reports statistical significance using exact McNemar tests and paired bootstrap tests, and reports confidence intervals and p-values for the main experimental results. See Section 3.4 and Table 7.

\item {\bf Experiments compute resources}
    \item[] Question: For each experiment, does the paper provide sufficient information on the computer resources (type of compute workers, memory, time of execution) needed to reproduce the experiments?
    \item[] Answer: \answerYes{} % Replace by \answerYes{}, \answerNo{}, or \answerNA{}.
    \item[] Justification: The paper reports the computational setting used for evaluation, including the hardware environment for retrieval experiments and the use of inference-only API-based language models as shown in Appendix H.
    
\item {\bf Code of ethics}
    \item[] Question: Does the research conducted in the paper conform, in every respect, with the NeurIPS Code of Ethics \url{https://neurips.cc/public/EthicsGuidelines}?
    \item[] Answer: \answerYes{} % Replace by \answerYes{}, \answerNo{}, or \answerNA{}.
    \item[] Justification: The work evaluates the robustness of language-based assistants using benchmark datasets and synthetic communication variants and does not involve harmful interventions or human experimentation. The authors believe the research conforms to the NeurIPS Code of Ethics.

\item {\bf Broader impacts}
    \item[] Question: Does the paper discuss both potential positive societal impacts and negative societal impacts of the work performed?
    \item[] Answer: \answerYes{} % Replace by \answerYes{}, \answerNo{}, or \answerNA{}.
    \item[] Justification: The paper discusses both positive and negative societal implications, including fair access to AI assistants across communication styles and English varieties, as well as reliability and equity concerns arising from system sensitivity to linguistic variation. See Discussion Sections 5.3 and 5.4.
    
\item {\bf Safeguards}
    \item[] Question: Does the paper describe safeguards that have been put in place for responsible release of data or models that have a high risk for misuse (e.g., pre-trained language models, image generators, or scraped datasets)?
    \item[] Answer: \answerNA{} % Replace by \answerYes{}, \answerNo{}, or \answerNA{}.
    \item[] Justification: The paper does not release a high-risk model, dataset, or system that requires special safeguards beyond standard research practices.

\item {\bf Licenses for existing assets}
    \item[] Question: Are the creators or original owners of assets (e.g., code, data, models), used in the paper, properly credited and are the license and terms of use explicitly mentioned and properly respected?
    \item[] Answer: \answerYes{} % Replace by \answerYes{}, \answerNo{}, or \answerNA{}.
    \item[] Justification: The paper identifies the datasets, benchmarks, models, and software assets used in the experiments, and cites their original sources.

\item {\bf New assets}
    \item[] Question: Are new assets introduced in the paper well documented and is the documentation provided alongside the assets?
    \item[] Answer: \answerNA{} % Replace by \answerYes{}, \answerNo{}, or \answerNA{}.
    \item[] Justification: No new dataset, model, or software asset is released as part of this submission.

\item {\bf Crowdsourcing and research with human subjects}
    \item[] Question: For crowdsourcing experiments and research with human subjects, does the paper include the full text of instructions given to participants and screenshots, if applicable, as well as details about compensation (if any)? 
    \item[] Answer: \answerNA{} % Replace by \answerYes{}, \answerNo{}, or \answerNA{}.
    \item[] Justification: The study does not involve crowdsourcing, human subjects, or human participants.

\item {\bf Institutional review board (IRB) approvals or equivalent for research with human subjects}
    \item[] Question: Does the paper describe potential risks incurred by study participants, whether such risks were disclosed to the subjects, and whether Institutional Review Board (IRB) approvals (or an equivalent approval/review based on the requirements of your country or institution) were obtained?
    \item[] Answer: \answerNA{} % Replace by \answerYes{}, \answerNo{}, or \answerNA{}.
    \item[] Justification: The study does not involve human subjects research and therefore does not require IRB review.

\item {\bf Declaration of LLM usage}
    \item[] Question: Does the paper describe the usage of LLMs if it is an important, original, or non-standard component of the core methods in this research? Note that if the LLM is used only for writing, editing, or formatting purposes and does \emph{not} impact the core methodology, scientific rigor, or originality of the research, declaration is not required.
    %this research? 
    \item[] Answer: \answerYes{} % Replace by \answerYes{}, \answerNo{}, or \answerNA{}.
    \item[] Justification: LLMs are a central component of the methodology and are used to generate communication-style variants, perform rubric-based validation, core experiments and grade benchmark outputs. Their usage is described in Taxonomy and Experiment Sections 2.2, 2.3, and 3.2.

\end{enumerate}

\end{document}